\documentclass{article}

\usepackage[preprint]{neurips_2026}

\usepackage[utf8]{inputenc} 
\usepackage[T1]{fontenc}    

\usepackage{hyperref}       
\usepackage{url}            
\usepackage{booktabs}       
\usepackage{amsfonts}       
\usepackage{nicefrac}       
\usepackage{microtype}      
\usepackage{xcolor}         

\usepackage{graphicx}
\usepackage{amsmath}
\usepackage{amssymb}
\usepackage{array}
\usepackage{multirow}
\usepackage{color}
\usepackage{colortbl}
\usepackage{framed}
\usepackage{bm}
\usepackage{xspace}
\usepackage{enumitem}
\usepackage{xparse}
\usepackage{algorithm}
\usepackage{listings}
\usepackage{url}
\usepackage{mathtools}
\usepackage{pifont}
\usepackage{bbm}

\definecolor{Gray}{gray}{0.2}
\definecolor{lightgray}{gray}{0.92}
\definecolor{blond}{rgb}{0.98, 0.94, 0.75}
\definecolor{TitleColor}{gray}{0.95}
\definecolor{LightCyan}{rgb}{0.88,0.95,1}
\definecolor{LighterGreen}{rgb}{0.94, 0.96, 0.93}
\definecolor{OurColor}{rgb}{0.90, 0.95, 0.90}
\definecolor{OurDarkColor}{rgb}{0.10, 0.55, 0.10}

\definecolor{blond}{rgb}{0.98, 0.94, 0.75}
\def \ie {\emph{i.e.}}
\def \eg {\emph{e.g.}}

\newcommand{\tit}[1]{\noindent\textbf{#1.}}
\newcommand{\tinytit}[1]{\noindent\textbf{#1.}}
\newcommand{\inc}[1]{\textcolor{OurDarkColor}{\textbf{\footnotesize$\Delta$+#1}}}
\newcommand{\dec}[1]{\textcolor{red}{\textbf{\footnotesize$\Delta$-#1}}}

\newcommand{\ours}{HeRA\xspace}

\newcommand{\llm}{\mathcal{G}}
\newcommand{\mllm}{\mathcal{M}}
\newcommand{\visenc}{\mathcal{V}}
\newcommand{\teachvisenc}{\mathcal{V}^{\textit{t}}}
\newcommand{\proj}{\mathrm{proj}}
\newcommand{\visembd}{\mathbf{v}}

\newcommand{\txtembd}{\mathbf{x}}
\newcommand{\losslm}{\mathcal{L}_{\text{LM}}}
\newcommand{\lossra}{\mathcal{L}_{\text{RA}}}
\newcommand{\losshera}{\mathcal{L}_{\text{HeRA}}}
\newcommand{\heads}{\mathcal{H}}
\newcommand{\topheads}{\mathcal{H}_m^{\text{top}}}
\newcommand{\worstheads}{\mathcal{H}_m^{\text{worst}}}
\newcommand{\mknn}{m_{k\mathrm{NN}}}
\newcommand{\dataset}{\mathcal{D}}
\newcommand{\batch}{\mathcal{B}}
\newcommand{\knnvis}{\mathcal{N}_k^{\teachvisenc}(I_i)}

\title{Mind the Heads: Topological Representation Alignment for Multimodal LLMs}

\author{%
  \textbf{Davide Caffagni}\textsuperscript{$\dag$1} \quad
  \textbf{Alberto Compagnoni}\textsuperscript{$\dag$1,2} \quad
  \textbf{Federico Melis}\textsuperscript{$\dag$1} \quad
  \textbf{Sara Sarto}\textsuperscript{1} \\
  \textbf{Pier Luigi Dovesi}\textsuperscript{3} \quad
  \textbf{Mark Granroth-Wilding}\textsuperscript{3} \quad
  \textbf{Marcella Cornia}\textsuperscript{1} \quad
  \textbf{Lorenzo Baraldi}\textsuperscript{1} \\
  \\
  \textsuperscript{1}University of Modena and Reggio Emilia \quad \textsuperscript{2}University of Pisa \quad \textsuperscript{3}AMD Silo AI
  \\[1ex]
  \tt\small\href{https://aimagelab.github.io/HeRA/}{\texttt{aimagelab.github.io/HeRA}}
}

\begin{document}

\maketitle

\begingroup
\renewcommand\thefootnote{$\dag$}
\footnotetext{Equal contribution. Emails: \texttt{\{name\}.\{surname\}@\{\textsuperscript{1}unimore.it, \textsuperscript{2}phd.unipi.it, \textsuperscript{3}amd.com\}}}
\endgroup

\vspace{-0.43cm}

\begin{abstract}
  Representation alignment has emerged as an effective approach to improve Multimodal Large Language Models (MLLMs) by regularizing their internal representations toward those of an external vision encoder. However, existing methods typically align a fixed layer of the language backbone, overlooking the fine-grained structure of Transformer models. In this work, we propose \textbf{\underline{He}}ad-Wise \textbf{\underline{R}}epresentation \textbf{\underline{A}}lignment (\textbf{\ours}), a method that enforces cross-modal alignment at the level of individual attention heads. Our approach is grounded in the Platonic Representation Hypothesis, focusing on preserving the \textit{topological structure} of representations (\ie, their local neighborhood relationships) across modalities. Following the Mutual K-Nearest Neighbor (MKNN) alignment metric, we introduce a contrastive objective that acts as a differentiable proxy for matching local structures. \ours applies this objective during multimodal training to specific attention heads in the LLM, selected by their alignment score according to the MKNN metric. Counterintuitively, we find that aligning the \textit{least} aligned heads yields the largest gains. Extensive evaluations across multiple MLLMs and 18 benchmarks demonstrate that \ours consistently improves performance on challenging vision-centric tasks and serves as an effective regularizer against visual hallucinations by naturally curbing the over-reliance on linguistic priors. Our code is publicly released.
\end{abstract}

\section{Introduction}
\label{sec:introduction}

Multimodal Large Language Models (MLLMs)~\cite{bai2025qwen3,liu2024improved,tong2024cambrian,wang2025internvl3} have emerged as powerful systems capable of solving a wide range of vision-language tasks. Despite their rapid progress, improvements are still largely driven by scaling data, model size, and post-training techniques, rather than by principled changes to their internal mechanisms. While current pipelines have proven highly effective for many applications, MLLMs still exhibit notable limitations in foundational visual reasoning scenarios. Tasks such as confirming the presence of specific objects, accurately counting them, understanding spatial relationships, or parsing dense visual information remain surprisingly challenging~\cite{fu2024blink,tong2024cambrian,tong2024eyes,wu2024v,grok}. This highlights a severe deficit in visual perception, raising a fundamental question: \textit{how can we improve multimodal reasoning by directly intervening on the interaction between vision and language within the model?}

A growing line of work addresses this deficiency through \textit{representation alignment}~\cite{caffagni2025seeing,wangreconstructive,yoon2025visual}: during multimodal training, the internal representations of the language model are regularized to match those of an external vision encoder. This can be interpreted as a form of cross-modal distillation, where the MLLM acts as a student and the vision encoder as a teacher, producing aligned representations of the same underlying content across modalities. While this technique has shown promise in improving visual grounding, existing approaches typically align a \textit{fixed} representation within the language backbone~\cite{yoon2025visual}, such as the middle layer, without accounting for the internal structure of the model.

This limitation is particularly relevant in MLLMs built upon pre-trained LLMs with strong language priors. Unlike diffusion-based models, where representation alignment is applied during training from scratch~\cite{leng2025repa,yurepresentation}, aligning representations in MLLMs may interact in complex ways with the pre-existing organization of the language model. In this setting, selecting \textit{which} representation to align becomes a critical design choice.

In this work, we pursue a more principled approach to representation selection, grounded in the \textit{Platonic Representation Hypothesis} (PRH)~\cite{groger2026revisiting,huh2024platonic}. PRH posits that representations learned across different modalities are \textit{locally consistent}: semantically similar inputs share the same neighborhood structure in their respective latent spaces. This can be interpreted as a form of \textit{topological alignment} across modalities, where the local geometry of the representation space is preserved, and can be quantified by the Mutual K-Nearest Neighbor (MKNN) metric, which measures the agreement between local neighborhoods. While prior work has established a positive correlation between MKNN alignment and downstream language performance~\cite{gan2025cross,huh2024platonic}, it remains unclear whether explicitly enforcing such alignment leads to improvements in MLLMs.

\begin{figure}[t]
    \centering
    \includegraphics[width=0.98\linewidth]{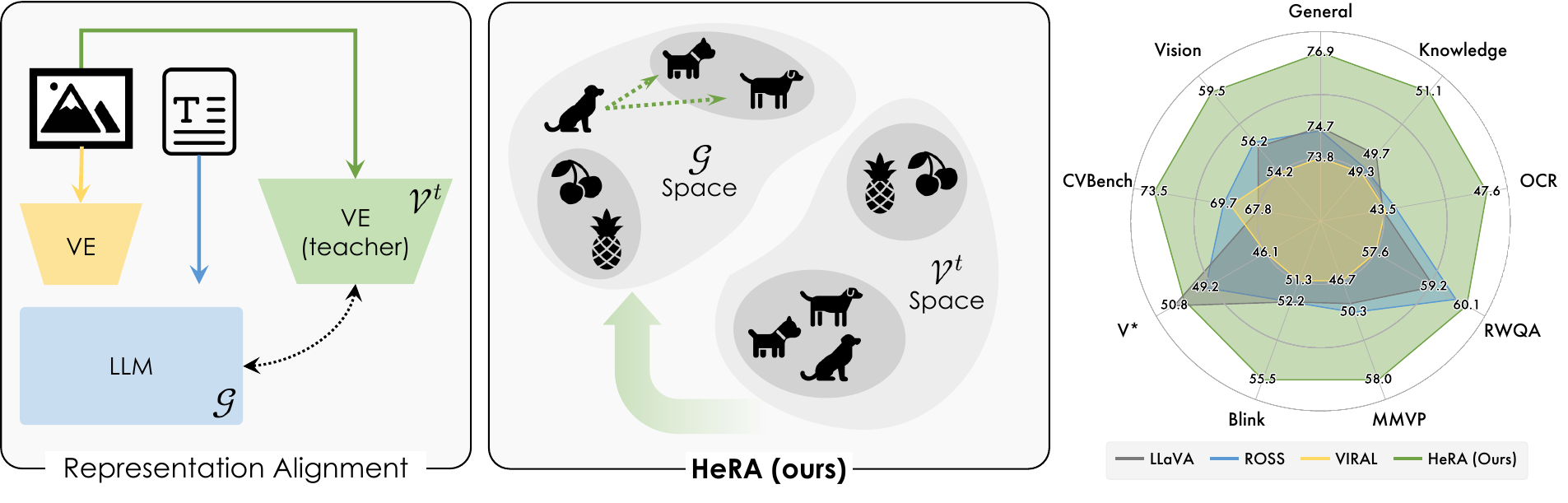}
    \vspace{-0.2cm}
    \caption{Standard representation alignment imposes strict vision-language feature matching (\textit{left}), while \ours (\textit{center}) matches cross-modal local neighbors, leading to superior VQA results (\textit{right}).}
    \label{fig:first_page}
    \vspace{-0.35cm}
\end{figure}

To address this, we propose \textbf{\underline{He}}ad-Wise \textbf{\underline{R}}epresentation \textbf{\underline{A}}lignment (\textbf{\ours}), a method that enforces cross-modal alignment at the level of individual attention heads rather than fixed, coarser layers. We use pre-computed MKNN scores as a diagnostic to guide this selection. Counterintuitively, we find that targeting the \textit{least} aligned heads yields the largest gains, as it strengthens misaligned components of the model while preserving already aligned structures. As outlined in Fig.~\ref{fig:first_page}, \ours applies a contrastive objective to these selected heads, encouraging their representations to match the \textit{local topological structure} induced by an external vision encoder. This serves as a differentiable proxy for MKNN alignment, promoting cross-modal consistency without imposing rigid feature matching that often conflicts with the language modeling objective.

We evaluate \ours across multiple MLLMs under the popular LLaVA~\cite{liu2024improved} framework. Extensive evaluations across 18 benchmarks~\cite{tong2024cambrian} demonstrate that \ours yields consistent improvements on challenging vision-centric tasks without sacrificing (and often improving) general visual question-answering performance. Furthermore, the topological alignment enforced by \ours serves as an effective regularizer against visual hallucinations~\cite{guan2024hallusionbench,wang2023amber}, naturally curbing the models' tendency to over-rely on linguistic priors.

\section{Related Work}
\label{sec:related}
\tinytit{The Platonic Representation Hypothesis}
The Platonic Representation Hypothesis (PRH)~\cite{huh2024platonic} posits that models trained across different architectures, modalities, and objectives converge toward structurally similar latent spaces. Crucially, concurrent work~\cite{groger2026revisiting} highlights that this structural consistency holds \textit{locally} rather than globally: semantically equivalent inputs preserve their neighborhood relationships across modalities, while the absolute global geometry may differ. While PRH shows that this local cross-modal alignment naturally emerges with scale and correlates with improved capabilities, it is unclear if a \textit{causal} relationship can be established. In this work, we investigate whether explicitly enforcing this local neighborhood consistency can lead to better MLLMs.

\tinytit{Vision-Centric Supervision in MLLMs}
Recent efforts to boost visual understanding in MLLMs have focused on introducing explicit vision-centric supervision. Several works attempt this by enforcing representation alignment between the MLLM and a teacher vision encoder. However, these methods typically operate directly on the visual features extracted from a fixed, hard-coded layer of the language backbone. For instance, JARVIS~\cite{caffagni2025seeing} reconstructs visual targets using representations from one-quarter of the LLM depth. VIRAL~\cite{yoon2025visual} aligns features from the middle layer, and ROSS~\cite{wangreconstructive} trains a denoiser using the final layer outputs. In contrast, \ours takes a fundamentally different approach: we enforce topological alignment within the \textit{textual} space of the MLLM (conditioned on the multimodal input) rather than strictly matching features from the vision teacher. Furthermore, we abandon the restrictive fixed-layer assumption entirely, instead targeting specific attention heads to preserve local neighborhood structures without conflicting with the language modeling task.

Research on MLLMs is moving fast~\cite{bai2025qwen3,wang2025internvl3}, thanks to massive datasets, post-training, stronger LLM backbones, and natively multimodal models~\cite{qwen35towards,tong2026beyond}. In this work, we study a novel representation alignment objective on the LLaVA~\cite{liu2024improved} framework to keep experiments computationally tractable, although we also apply it on top of state-of-the-art LLMs, such as the latest Qwen3~\cite{yang2025qwen3} family.
\section{Proposed Method}
\label{sec:method}

\subsection{Background}
\begin{figure}[t]
    \centering
    \includegraphics[width=0.98\linewidth]{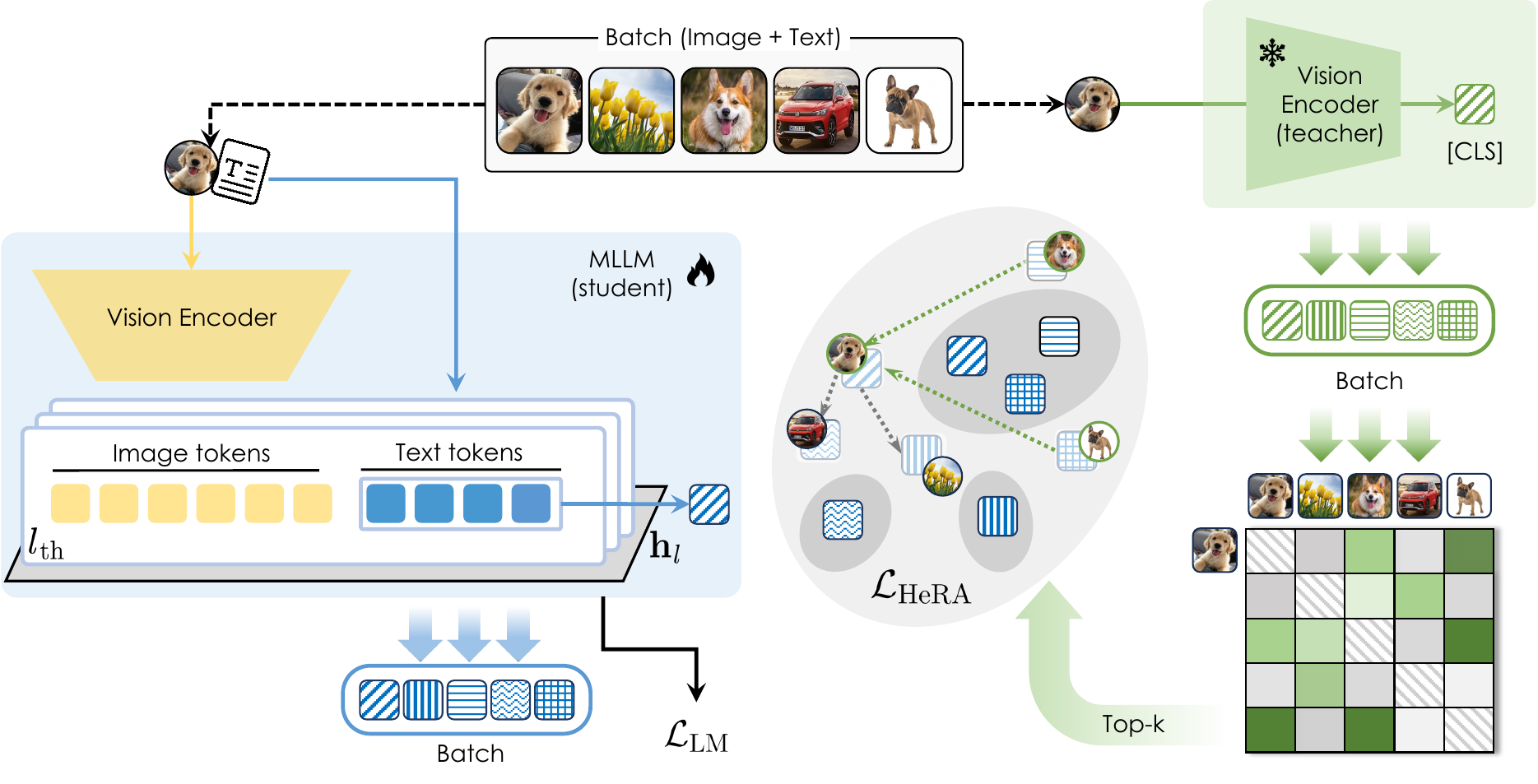}
    \vspace{-0.2cm}
    \caption{\textbf{Overview of \ours.} Alongside the standard language modeling objective ($\mathcal{L}_{\text{LM}}$), \ours employs a contrastive loss ($\mathcal{L}_{\text{HeRA}}$) to pull representations from selected LLM attention heads closer to their $k$-nearest neighbors (Top-$k$), computed in the latent space of a frozen teacher vision encoder.}    
    \label{fig:method}
    \vspace{-0.35cm}
\end{figure}

\tinytit{Multimodal Large Language Models (MLLMs)}
From an architectural perspective (refer to Fig.~\ref{fig:method}, \textit{left}), an MLLM $\mllm$ comprises \textit{(i)} an LLM $\llm$, which constitutes the reasoning backbone and natural language interface of the model; \textit{(ii)} a pre-trained vision encoder $\visenc$ to process visual inputs; and \textit{(iii)} a projector $\proj$, which aligns the output embedding space of $\visenc$ with the input embedding space of $\llm$.

$\mllm$ ingests and generates text $x$ as a sequence of tokens $\txtembd_{1,\dots,T}$ converted into latent vectors by the embedding matrix of $\llm$. On the other hand, visual inputs $I$ are first processed by the vision encoder $\visenc$, then converted into the input embedding space of $\llm$ by the projector: $\visembd_{1,\dots,V} = \proj(\visenc(I))$,
and finally concatenated to the sequence of text embeddings. We train $\mllm$ to minimize the negative log-likelihood of generating token $\txtembd_{j}$ given the image $I$ and the preceding text $\txtembd_{1,\dots,j-1}$: 
\begin{equation}
    \label{eq:loss_lm}
    \losslm(I_i, x_i, \mllm) = - \sum_{j}^{T} \log P\left(\txtembd_j | \visembd_{1,\dots,V}, \txtembd_{1,\dots,j-1} ; \mllm\right).
\end{equation}

\tinytit{Mutual K-Nearest Neighbor (MKNN) Alignment Metric}
MKNN~\cite{huh2024platonic} is a kernel alignment metric enabling comparison between different representation functions. In this work, we measure the alignment between textual and visual representations of the same data point. Given an image-text pair $(I, x)_i \in \dataset$, where $\dataset$ is a dataset of aligned image-text pairs, we denote by $\llm(x_i) \in \mathbb{R}^{d_{\llm}}$ a representation of the text extracted from the language model, and by $\teachvisenc(I_i) \in \mathbb{R}^{d_{\teachvisenc}}$ representation of the corresponding image from a \textit{teacher} vision encoder. Here, $\llm(x_i)$ refers to an internal representation (\eg, from intermediate layers or attention heads).

For a dataset $\dataset$, MKNN measures the agreement between the local neighborhood structures induced by the two representation spaces, by computing the average intersection of their $k$-nearest neighbor sets. We denote by $\mathcal{N}_k^{\mathcal{F}}(\cdot)$ the operator returning the $k$-nearest neighbors according to maximum dot product similarity in the latent space of the embedding function $\mathcal{F}$. For instance, in the language space, where we average pool the output embeddings, it is formally defined as follows:

\begin{equation}
    \mathcal{N}_k^{\llm}(x_i) = \text{argmax}_{j \neq i}^{(k)} \llm(x_i)^\top \llm(x_j).
\end{equation}
In the visual space, we pool by taking the \texttt{CLS} embeddings at the output of the vision encoder $\teachvisenc$. The MKNN alignment metric between $\llm$ and $\teachvisenc$ is thus defined as:

\begin{equation}
    \label{eq:mknn}
    \mknn(\llm, \teachvisenc, \dataset) = \mathop{\mathbb{E}}_{(I, x)_i \in \dataset} \left[ \frac{1}{k} \left| \mathcal{N}_k^{\llm}(x_i) \cap \mathcal{N}_k^{\teachvisenc}(I_i) \right| \right] \in [0, 1],
\end{equation}
where $|\cdot|$ denotes set cardinality. 

High scores in Eq.~\ref{eq:mknn} reflect that the \textit{local} topological latent structure generated by $\llm$ is preserved in the latent space of $\teachvisenc$.

\subsection{Contrastive Learning as a Proxy for Representation Alignment}
For a fixed vision encoder $\teachvisenc$, $\mknn$ has been positively correlated with better performance on language modeling tasks~\cite{huh2024platonic}. We want to probe whether a causal effect could exist in a multimodal scenario: \textit{can we train a better MLLM by explicitly enforcing alignment with the visual domain?} 

A natural approach would be to directly maximize $\mknn$ during training. However, Eq.~\ref{eq:mknn} depends on discrete neighbor indices and is therefore not differentiable. To address this, we propose a contrastive objective that encourages the multimodal representations produced by $\mllm$ to match the local neighborhood structure induced by the teacher vision encoder.

Given a batch $\batch = \{(I, x)_i\}$, let $\mllm(I_i, x_i) \in \mathbb{R}^{d_{\llm}}$ denote the multimodal representation obtained by average pooling the text embeddings\footnote{As text embeddings are conditioned on visual inputs, they can be considered inherently multimodal.}. For each sample $i$, we first identify a set of target neighbors $\knnvis$, corresponding to the $k$ nearest neighbors of $I_i$ in the teacher vision space. We then train $\mllm$ so that its representation $\mllm(I_i, x_i)$ is close to the representations of these neighbors, while being separated from the rest of the batch. Formally, this can be achieved via a multi-target variant of the InfoNCE~\cite{oord2018representation} loss:
\begin{equation}
\label{eq:loss_ra}
    \lossra(I_i, x_i, \mllm) = - \frac{1}{k} \sum_{j \in \knnvis} \log 
    \frac{\exp\left(\frac{\mllm(I_i, x_i)^\top \mllm(I_j, x_j)}{\tau}\right)}
    {\sum_{z \in \batch, z \neq i} \exp\left(\frac{\mllm(I_i, x_i)^\top \mllm(I_z, x_z)}{\tau}\right)},
\end{equation}
where $\tau$ is a learnable scalar governing the sharpness of the distribution. Minimizing Eq.~\ref{eq:loss_ra} teaches the \textit{student} model $\mllm$ to produce multimodal representations sharing the same \textit{local} neighborhood as the corresponding visual representations from the \textit{teacher} model $\teachvisenc$, which is exactly the property measured by the $\mknn$ metric.

\subsection{Head-Wise Representation Alignment (\ours)}
While the contrastive objective in Eq.~\ref{eq:loss_ra} enforces alignment at the level of a single pooled representation, it does not account for the internal structure of the language backbone. In particular, $\llm$ processes the entire multimodal sequence, suggesting that alignment can be more precisely controlled by operating directly on its internal representations.

In principle, $\llm$ generates multiple representations for a given input. Indeed, we can collect a representation from each Transformer layer of the language backbone. For language modeling (\ie, Eq.~\ref{eq:loss_lm}), we care about the last layer to sample the next token (after passing through the unembedding matrix). Conversely, representation alignment methods~\cite{leng2025repa,yurepresentation} typically rely on intermediate layers. However, the choice of which layer(s) to use is typically treated as a fixed hyperparameter (\eg, selecting the middle layer~\cite{yoon2025visual}), which does not adapt to the specific structure of a given model.

In this work, we instead probe finer-grained representations within the language model, specifically focusing on the individual attention heads in each multi-head self-attention layer of $\llm$. Because different attention heads specialize in different roles within an LLM~\cite{namcausal,olsson2022context,wanginterpretability}, working at the head level enables a more atomic intervention on the language model, mitigating potential conflicting effects between language modeling and representation alignment.

\tit{Head-level Representations}
In standard multi-head attention, the final output of a layer is obtained by concatenating the outputs of the individual heads and multiplying them by an output projection matrix $\mathbf{W}_O \in \mathbb{R}^{d_{\llm} \times d_{\llm}}$.
Let $\mathbf{h}_{l,h} \in \mathbb{R}^{d_{head}}$ be the output of the $h$-th attention head in layer $l$, where $d_{head} = d_{\llm}/H$. The output projection can be written as:
\begin{equation}
    \text{MultiHead}(\cdot) = [\mathbf{h}_{l,1}(\cdot), \dots, \mathbf{h}_{l,H}(\cdot)] \mathbf{W}_O.
\end{equation} 
Because matrix multiplication is a linear operator, we can decompose $\mathbf{W}_O$ into $H$ distinct blocks along its row dimension, such that $\mathbf{W}_O = [\mathbf{W}_{O,1}^\top, \dots, \mathbf{W}_{O,H}^\top]^\top$, with each $\mathbf{W}_{O,h} \in \mathbb{R}^{d_{head} \times d_{\llm}}$.
The multi-head attention output can then be equivalently written as:
\begin{equation}
    \text{MultiHead}(\cdot) = \sum_{h=1}^H \mathbf{h}_{l,h}(\cdot) \mathbf{W}_{O,h}.
\end{equation}

This decomposition allows us to isolate the projected contribution of each head before it is summed into the shared residual stream. Given a multimodal input $(I_i,x_i)$, we define:
\begin{equation}
    \mllm^{l,h}(I_i,x_i) = \mathbf{h}_{l,h}(I_i,x_i)\mathbf{W}_{O,h} \in \mathbb{R}^{d_{\llm}},
\end{equation}
where $\mllm^{l,h}$ denotes the representation extracted from the $h$-th attentive head in the $l$-th layer of $\mllm$ during the multimodal forward pass.

\begin{figure}[t]
    \centering
    \includegraphics[width=\linewidth]{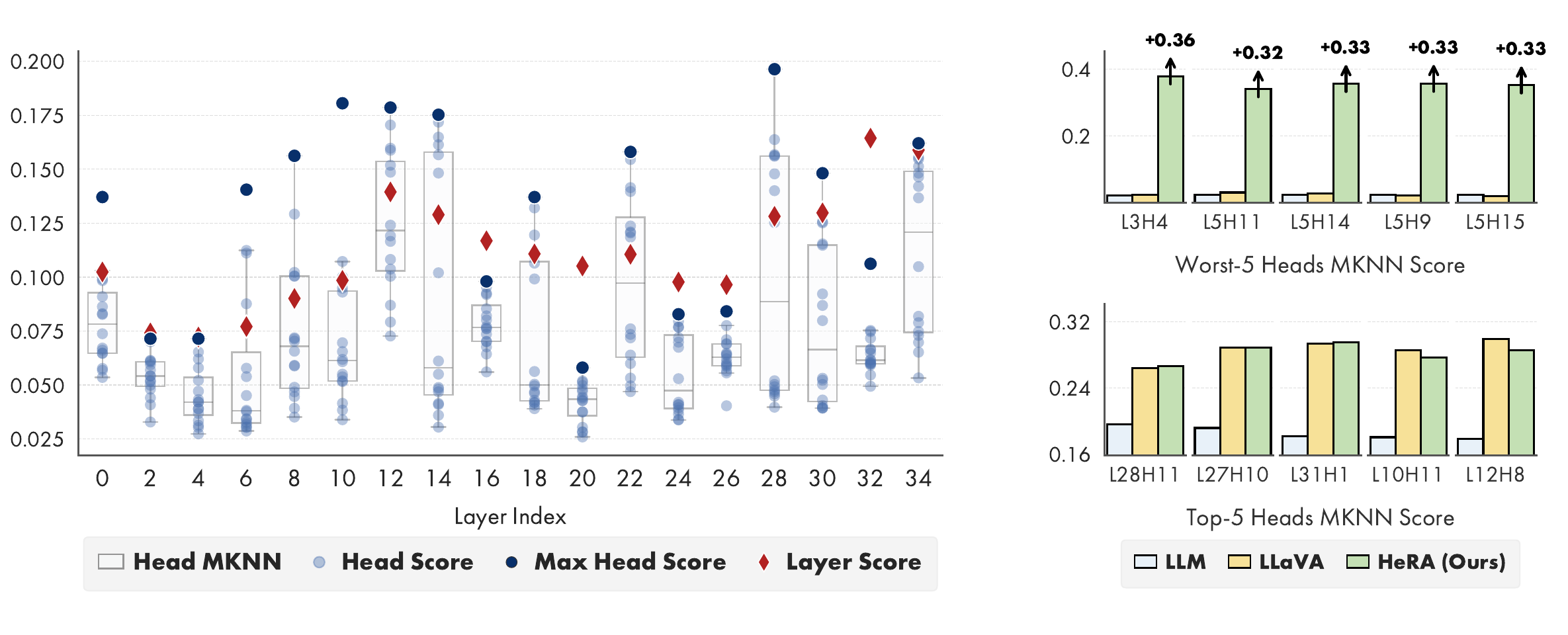}
    \vspace{-0.8cm} 
    \caption{\textbf{Left:} Alignment with DINOv2-L, measured with the MKNN metric on each layer and attention head of Qwen2.5-3B. \textbf{Right:} MKNN scores of the Worst-5 and Top-5 heads, computed on \textit{(i)} the base LLM; \textit{(ii)} after the LLaVA multimodal training; and \textit{(iii)} after the addition of \ours.}      
    \label{fig:mknn_plot}
    \vspace{-0.35cm}
\end{figure}

\tit{Head-wise Alignment Objective}
We apply the contrastive alignment loss of Eq.~\ref{eq:loss_ra} independently to selected head-level representations. For a set of layer-head indices $\heads = \{(l,h)\}$, we define our head-wise representation alignment loss as the average alignment loss over the selected heads:
\begin{equation}
    \label{eq:loss_hera}
    \losshera(I_i, x_i, \mllm, \heads) =
    \mathop{\mathbb{E}}_{(l,h)\in \heads}
    \left[
        \lossra(I_i, x_i, \mllm^{l,h})
    \right].
\end{equation}

The final training objective (illustrated in Fig.~\ref{fig:method}) is given by the sum of the language modeling and head-wise representation alignment losses:
\begin{equation}
    \label{eq:loss}
    \mathcal{L}(I_i, x_i, \mllm, \heads) = \losslm(I_i, x_i, \mllm) + \lambda \losshera(I_i, x_i, \mllm, \heads),
\end{equation}
where $\lambda$ is a fixed hyperparameter to balance the two contributions.

\tinytit{Heads Selection}
For an LLM with $L$ layers and $H$ attention heads, the total number of heads is $L \times H$. In practice, this number is in the order of hundreds, making an extensive search for the optimal $\heads$ unfeasible. To this end, we propose to exploit the $\mknn$ metric of Eq.~\ref{eq:mknn} to rank the heads by their alignment score with the vision encoder $\teachvisenc$. We compute this rank using the language model $\llm$ \textit{before} the multimodal training, so that its representations are purely textual. Surprisingly, we find that there always exists a set of heads whose alignment score greatly exceeds that of any layer in the same model (see Fig.~\ref{fig:mknn_plot}, left). Once the alignment rank is computed, we posit to select a subset of $m$ heads following two reasonable strategies. Specifically, we select either \textit{(i)} the best aligned heads (\ie, $\topheads$), as forcing the alignment should be easier because they start from an already partially aligned latent space, or \textit{(ii)} the least aligned heads (\ie, $\worstheads$), so to strengthen the components of the model further away from the visual domain. Empirically, we find that choosing $\worstheads$ works best: it boosts the alignment of poorly aligned heads, while preserving the alignment of the strongest heads, as displayed in Fig.~\ref{fig:mknn_plot}, right.

\section{Experiments}
\label{sec:experiments}

\subsection{Experimental Settings}
\label{sec:settings}

\tit{Training Details} We train all models following the two-stage LLaVA-1.5 pipeline~\cite{liu2024improved}, with the same training data and protocol. As vision encoder $\visenc$, we adopt SigLIP2 ViT-SO400M/14@384~\cite{tschannen2025siglip} across all experiments. For the LLM $\llm$, we consider a diverse set of architectures, including Vicuna~\cite{vicuna2023}, LLama3~\cite{grattafiori2024llama}, Qwen2.5~\cite{qwen2024qwen25technicalreport}, and Qwen3~\cite{yang2025qwen3}, ranging from 3B to  14B parameters. We apply the \ours loss (cf. Eq.~\ref{eq:loss_hera}) in both training stages, using $\lambda$ equal to 0.01, and $k$ equal to 10. Unless otherwise specified, we use DINOv2 ViT-L~\cite{oquab2024dinov2} as teacher vision encoder $\teachvisenc$.

\tit{Head Selection} We perform head selection \emph{prior} to multimodal training. Specifically, we compute the $\mknn$ alignment score (Eq.~\ref{eq:mknn}) for each head using the LLM $\llm$, before any multimodal finetuning. This produces a ranking of heads based on their degree of alignment with the visual domain. The scores are computed on 1{,}000 samples from the GranD dataset~\cite{hanoona2023GLaMM}, which provides highly detailed captions enabling a reliable estimation of cross-modal neighborhood structure. We select the $m=5$ least aligned heads and restrict the application of \ours to this subset throughout training. 

\begin{table}[t]
\caption{Ablation study on the choice of \textit{(i)} the objective for representation alignment (feature- vs. contrastive-based), and \textit{(ii)} the granularity of the LLM representation to align (layer- vs. head-level).}
\label{tab:ablation}
\vspace{-0.1cm}
\centering
\setlength{\tabcolsep}{0.3em}
\resizebox{\linewidth}{!}{
\begin{tabular}{ccc c ccccc c ccccc}
\toprule
\multicolumn{3}{c}{\textbf{Representation Alignment}} & & \multicolumn{5}{c}{\textbf{LLM:} Qwen2.5-3B} & & \multicolumn{5}{c}{\textbf{LLM:} Qwen3-4B} \\
\cmidrule{1-3} \cmidrule{5-9} \cmidrule{11-15}
Objective & Granularity & Selection & & \textbf{General} & \textbf{Knowledge} & \textbf{OCR} & \textbf{Vision} & \textbf{All} & & \textbf{General} & \textbf{Knowledge} & \textbf{OCR} & \textbf{Vision} & \textbf{All} \\
\midrule
- & - & - & & 73.5 & 46.7 & 42.2 & 50.5 & 54.2 & & 75.6 & 49.6 & 43.8 & 56.3 & 57.4 \\
Feature & Layer & Middle & & 72.6 & 46.4 & 42.3 & 50.1 & 53.8 & & 75.0 & 48.6 & 42.8 & 53.4 & 56.0 \\
Feature & Head & Worst (5) & & 74.0 & 46.5 & 42.8 & 51.6 & 54.7 & & 76.0 & 49.3 & 44.7 & 55.9 & 57.6 \\
\midrule
Contrastive & Layer & Middle & & 72.8 & 45.7 & 41.6 & 51.1 & 53.8 & & 75.8 & 49.5 & 43.7 & 57.2 & 57.6 \\
Contrastive & Layer & Worst (5) & & 73.1 & 46.2 & 40.9 & 51.7 & 54.0 & & 73.2 & 47.9 & 41.2 & 52.1 & 54.6 \\
\midrule
Contrastive & Head & Random (5) & & 73.7 & 46.6 & 42.1 & 49.7 & 54.0 & & \textbf{76.1} & 49.7 & 44.0 & 56.6 & 57.7 \\
Contrastive & Head & Top (5) & & 73.8 & 46.5 & 42.8 & 50.2 & 54.3 & & 75.9 & 49.9 & \textbf{45.2} & 56.3 & 57.8 \\
\midrule
Contrastive & Head & Worst (1) & & 73.5 & 46.9 & 42.9 & 51.2 & 54.6 & & 76.0 & 49.7 & 44.1 & 57.0 & 57.8 \\
Contrastive & Head & Worst (3) & & 73.9 & \textbf{47.6} & 43.6 & 51.0 & 55.0 & & 75.8 & 49.9 & 44.7 & 57.0 & 57.9 \\
Contrastive & Head & Worst (10) & & 62.6 & 45.5 & 34.6 & 38.9 & 46.0 & & 75.8 & 49.7 & 44.7 & 56.3 & 57.7 \\
\rowcolor{OurColor}
\textbf{Contrastive} & \textbf{Head} & \textbf{Worst (5)} & & \textbf{74.5} & 47.5 & \textbf{43.8} & \textbf{52.9} & \textbf{55.7} & & 76.0 & \textbf{50.1} & 44.5 & \textbf{58.5} & \textbf{58.4} \\
\bottomrule
\end{tabular}
}
\vspace{-0.35cm}
\end{table}

\tit{Evaluation Benchmarks} 
We primarily evaluate our method using the Cambrian comprehensive benchmark suite~\cite{tong2024cambrian} covering General, Knowledge, OCR, and Vision tasks. 
In addition, we evaluate hallucination robustness on CHAIR-MSCOCO~\cite{yue2024less}, AMBER~\cite{wang2023amber}, and HallusionBench~\cite{guan2024hallusionbench}. The complete details on the evaluation datasets are reported in Appendix~\ref{sec:appendix_benchmarks}.

\subsection{Ablation Studies and Analyses} 
We start by presenting a set of ablation studies in Table~\ref{tab:ablation} designed to understand how key architectural and objective choices influence the behavior of \ours. For these experiments, we employ Qwen2.5-3B~\cite{qwen2024qwen25technicalreport} and Qwen3-4B~\cite{yang2025qwen3} as the underlying LLMs. As a baseline, we consider a LLaVA model trained on top of the same LLMs without alignment regularization (\textit{first} row). For all configurations, we use the same training settings used in our approach, as described in Sec.~\ref{sec:settings}.

\tinytit{Objective and Granularity} 
First, we consider a standard representation alignment approach~\cite{yoon2025visual,yurepresentation}, where the LLM is trained to minimize the cosine similarity between visual-token features at the middle layer and those from the teacher vision encoder (\textit{second} row), using a trainable projector. While this is ineffective on both LLMs, switching to our contrastive learning objective using the same representation granularity (\textit{fourth} row) shows promising results on vision tasks. 

\tinytit{Head-Level Alignment and Selection} 
Sticking with the contrastive objective, we move to a finer granularity, considering the textual representations from specific attentive heads in the LLM. The head selection criterion follows the MKNN alignment score with the vision encoder: we select either the top-5 (\textit{seventh} row) or worst-5 (\textit{last} row) according to this ranking. On both LLMs, we record a striking difference favoring alignment on the worst-5 heads. For instance, Qwen2.5-3B boosts its performance on vision-centric tasks by +1.4 points, whereas Qwen3-4B enjoys a +2.3 points gain. As control trials, we also apply the contrastive alignment on a random subset of 5 heads (\textit{sixth} row), yielding no clear benefit, and experiment with the ``worst-5'' selection criterion at the layer-level (\textit{fifth} row), which actually registers a mild performance regression on Qwen2.5-3B and a severe degradation on Qwen3-4B.

\tinytit{Connection to the Platonic Representation Hypothesis}
The superiority of the worst-5 strategy corroborates the positive correlation between alignment and performance reported by the PRH~\cite{huh2024platonic}. As shown in Fig.~\ref{fig:mknn_plot} (\textit{right}), after \ours training, the worst-5 heads drastically increase their vision-language alignment without penalizing the alignment of the top-5 heads. Conversely, explicitly forcing alignment on the top-5 heads has no meaningful collateral impact on the worst-5 heads. Interestingly, aligning the visual features from the worst-5 heads (\textit{third} row) is ineffective, and has little impact on their alignment scores (see the plot in Fig.~\ref{fig:mknn_top_supp} in Appendix~\ref{sec:appendix_results}).

\tinytit{Number of Heads to Align}
Finally, we ablate the number of heads to align (Table~\ref{tab:ablation}, \textit{bottom}). Using 3 heads, or even a single one, yields modest gains on both LLMs, while the best overall performance is achieved with 5 heads, particularly on challenging vision benchmarks. Conversely, further scaling up the number of heads to 10 leads to a regression, particularly with Qwen2.5-3B, indicating that aligning too many heads begins to conflict with the core language modeling task.

\begin{table}[t]
\centering
\caption{VQA results of \ours applied to the LLaVA training recipe on different LLMs.}
\label{tab:llms}
\vspace{-0.1cm}
\setlength{\tabcolsep}{.42em}
\resizebox{0.92\linewidth}{!}{%
\begin{tabular}{l c c c c c c c >{\columncolor{OurColor!30}}c>{\columncolor{OurColor!30}}c>{\columncolor{OurColor!30}}c>{\columncolor{OurColor!30}}c>{\columncolor{OurColor!30}}c>{\columncolor{OurColor!30}}c>{\columncolor{OurColor!30}}c}
\toprule 
& & \textbf{General} & & \textbf{Knowledge} & & \textbf{OCR} & & \multicolumn{7}{c}{\cellcolor{OurColor!30}\textbf{Vision}} \\
\cmidrule(lr){3-3} \cmidrule(lr){5-5} \cmidrule(lr){7-7} \cmidrule{9-15}
\textbf{Model} & & Avg & & Avg & & Avg & & RWQA & MMVP & Blink & V* & CVBench & Avg & \\
\midrule
Qwen2.5-3B & & 73.5 & & 46.7 & & 42.2 & & 55.2 & 46.0 & 46.8 & 44.5 & \textbf{60.2} & 50.5 & \\
\rowcolor{OurColor}
\textbf{+ \ours (Ours)} & & \textbf{74.5} & & \textbf{47.5} & & \textbf{43.8} & & \textbf{56.3} & \textbf{48.0} & \textbf{49.1} & \textbf{51.3} & 59.6 & \textbf{52.9} & \inc{2.4} \\
\midrule
Qwen3-4B & & 75.6 & & 49.6 & & 43.8 & & 59.9 & 48.0 & \textbf{55.1} & 50.3 & 68.3 & 56.3 & \\
\rowcolor{OurColor}
\textbf{+ \ours (Ours)} & & \textbf{76.0} & & \textbf{50.1} & & \textbf{44.5} & & \textbf{61.3} & \textbf{56.0} & 53.7 & \textbf{52.4} & \textbf{69.1} & \textbf{58.5} & \inc{2.2} \\
\midrule
Vicuna-7B & & \textbf{72.2} & & 44.3 & & \textbf{45.7} & & 56.5 & 38.7 & 46.8 & 44.5 & \textbf{62.1} & 49.7 & \\
\rowcolor{OurColor}
\textbf{+ \ours (Ours)} & & 72.1 & & \textbf{44.5} & & \textbf{45.7} & & \textbf{57.8} & \textbf{42.7} & \textbf{47.9} & \textbf{49.7} & 61.9 & \textbf{52.0} & \inc{2.3} \\
\midrule
LLama3-8B & & 73.3 & & 45.0 & & 43.0 & & 60.1 & 46.0 & 49.2 & 44.0 & \textbf{69.5} & 53.8 & \\
\rowcolor{OurColor}
\textbf{+ \ours (Ours)} & & \textbf{74.6} & & \textbf{46.3} & & \textbf{44.7} & & \textbf{60.4} & \textbf{46.7} & \textbf{50.2} & \textbf{51.8} & 66.4 & \textbf{55.1} & \inc{1.3} \\
\midrule
Qwen2.5-7B & & 76.2 & & 50.2 & & 47.9 & & 59.6 & 51.3 & \textbf{51.7} & \textbf{50.8} & 70.2 & 56.7 & \\
\rowcolor{OurColor}
\textbf{+ \ours (Ours)} & & \textbf{76.5} & & \textbf{50.5} & & \textbf{48.6} & & \textbf{61.3} & \textbf{54.0} & 50.2 & 50.3 & \textbf{71.3} & \textbf{57.4} & \inc{0.7} \\
\midrule
Qwen3-8B & & 74.7 & & 49.7 & & 43.5 & & 59.2 & 49.3 & 52.2 & \textbf{50.8} & 67.8 & 55.9 & \\
\rowcolor{OurColor}
\textbf{+ \ours (Ours)} & & \textbf{76.9} & & \textbf{51.1} & & \textbf{47.6} & & \textbf{60.1} & \textbf{58.0} & \textbf{55.5} & 50.3 & \textbf{73.5} & \textbf{59.5} & \inc{3.6} \\
\midrule
Vicuna-13B & & 73.4 & & 45.5 & & \textbf{47.7} & & \textbf{58.7} & \textbf{44.7} & 50.6 & 46.1 & 63.7 & 52.7 & \\
\rowcolor{OurColor}
\textbf{+ \ours (Ours)} & & \textbf{73.6} & & \textbf{45.7} & & 47.6 & & 57.3 & 44.0 & \textbf{52.3} & \textbf{49.2} & \textbf{66.5} & \textbf{53.9} & \inc{1.2} \\
\midrule
Qwen2.5-14B & & 75.6 & & 50.7 & & 44.8 & & \textbf{60.1} & 47.3 & 52.7 & 46.6 & 67.9 & 54.9 & \\
\rowcolor{OurColor}
\textbf{+ \ours (Ours)} & & \textbf{77.4} & & \textbf{52.8} & & \textbf{49.3} & & \textbf{60.1} & \textbf{52.0} & \textbf{55.2} & \textbf{52.4} & \textbf{71.6} & \textbf{58.3} & \inc{3.4} \\
\midrule
Qwen3-14B & & 77.4 & & \textbf{52.8} & & 46.1 & & 60.3 & 57.3 & \textbf{52.6} & 50.3 & \textbf{70.8} & 58.2 & \\
\rowcolor{OurColor}
\textbf{+ \ours (Ours)} & & \textbf{77.7} & & 52.6 & & \textbf{47.8} & & \textbf{62.5} & \textbf{58.0} & 52.0 & \textbf{51.8} & 70.2 & \textbf{58.9} & \inc{0.7} \\
\bottomrule
\end{tabular}
}
\vspace{-0.3cm}
\end{table}

\begin{table}[t]
\caption{Results of \ours on visual hallucinations benchmarks.}
\label{tab:hall}
\vspace{-0.1cm}
\centering
\setlength{\tabcolsep}{0.35em}
\resizebox{\linewidth}{!}{
\begin{tabular}{lc cc c cccc c cccc c ccc}
\toprule
& & \multicolumn{2}{c}{\textbf{MSCOCO}}
& & \multicolumn{4}{c}{\textbf{AMBER (Generative)}}
& & \multicolumn{4}{c}{\textbf{AMBER (Discriminative)}}
& & \multicolumn{3}{c}{\textbf{HallusionBench}} \\
\cmidrule{3-4} \cmidrule{6-9} \cmidrule{11-14} \cmidrule{16-18}
& & CHAIR$_s$ $\downarrow$ & CHAIR$_i$ $\downarrow$ & & CHAIR$_i$ $\downarrow$ & Cover $\uparrow$ & HalRate $\downarrow$ & Cog $\downarrow$ & &  Acc $\uparrow$ & Prec $\uparrow$ & Rec $\uparrow$ & F1 $\uparrow$ & & qAcc $\uparrow$ & Easy $\uparrow$ & Hard $\uparrow$ \\
\midrule
Qwen2.5-3B & & 44.2 & 12.6 & & 5.8 & 52.0 & 30.1 & 3.3 & & 83.6 & 83.9 & \textbf{93.2} & 88.3 & & 23.7 & 60.7 & 55.2 \\
\rowcolor{OurColor}
\textbf{+ \ours (Ours)} & & \textbf{44.0} & \textbf{11.8} & & \textbf{5.2} & \textbf{52.3} & \textbf{27.6} & \textbf{2.8} & & \textbf{85.6} & \textbf{86.5} & 92.8 & \textbf{89.5} & & \textbf{24.4} & \textbf{61.3} & \textbf{60.2} \\
\midrule
Qwen3-4B & & \textbf{42.0} & \textbf{11.7} & & \textbf{5.5} & \textbf{53.0} & 28.2 & 3.0 & & \textbf{86.5} & 89.9 & \textbf{89.6} & \textbf{89.7} & & \textbf{22.2} & \textbf{63.1} & \textbf{54.6} \\
\rowcolor{OurColor}
\textbf{+ \ours (Ours)} & & 43.8 & 11.9 & & \textbf{5.5} & 52.5 & \textbf{27.6} & \textbf{2.8} & & \textbf{86.5} & \textbf{90.1} & 89.4 & \textbf{89.7} & & 18.9 & 60.7 & 50.3 \\
\midrule
Vicuna-7B & & \textbf{46.6} & \textbf{12.6} & & 6.3 & 52.2 & \textbf{30.8} & \textbf{3.3} & & 84.2 & 90.0 & 85.8 & 87.8 & & 15.6 & \textbf{55.4} & 45.7 \\
\rowcolor{OurColor}
\textbf{+ \ours (Ours)} & & 47.6 & 12.9 & & \textbf{6.1} & \textbf{52.6} & 31.9 & 3.5 & & \textbf{85.9} & \textbf{90.8} & \textbf{87.7} & \textbf{89.2} & & \textbf{16.3} & 55.0 & \textbf{49.6} \\
\midrule
LLama3-8B & & 44.8 & 13.0 & & \textbf{5.5} & 51.5 & 27.5 & 2.9 & & \textbf{85.9} & \textbf{89.2} & 89.6 & 89.4 & & 16.9 & \textbf{58.7} & \textbf{49.9} \\
\rowcolor{OurColor}
\textbf{+ \ours (Ours)} & & \textbf{40.2} & \textbf{11.7} & & 5.6 & \textbf{51.7} & \textbf{26.6} & \textbf{2.7} & & \textbf{85.9} & 88.3 & \textbf{90.8} & \textbf{89.5} & & \textbf{17.4} & 58.0 & 49.3 \\
\midrule
Qwen2.5-7B & & 45.2 & \textbf{11.9} & & 5.3 & 52.3 & 26.7 & \textbf{2.7} & & 87.6 & \textbf{89.6} & 91.9 & 90.7 & & 21.5 & \textbf{66.8} & 49.4 \\
\rowcolor{OurColor}
\textbf{+ \ours (Ours)} & & \textbf{44.8} & 12.4 & & \textbf{4.9} & \textbf{52.5} & \textbf{25.1} & \textbf{2.7} & & \textbf{87.7} & 89.1 & \textbf{92.8} & \textbf{90.9} & & \textbf{23.3} & 66.6 & \textbf{50.9} \\
\midrule
Qwen3-8B & & 42.0 & 12.3 & & 5.8 & 52.8 & 28.9 & 3.2 & & 87.3 & 90.3 & 90.6 & 90.4 & & 21.1 & 61.1 & 53.3 \\
\rowcolor{OurColor}
\textbf{+ \ours (Ours)} & & \textbf{39.2} & \textbf{11.0} & & \textbf{5.2} & \textbf{53.1} & \textbf{27.9} & \textbf{3.0} & & \textbf{89.0} & \textbf{92.4} & \textbf{90.9} & \textbf{91.6} & & \textbf{25.7} & \textbf{66.2} & \textbf{54.5} \\
\midrule
Vicuna-13B & & \textbf{43.0} & \textbf{11.9} & & \textbf{6.1} & 52.5 & \textbf{28.4} & 3.3 & & 84.8 & \textbf{93.9} & 82.4 & 87.8 & & 13.0 & 55.2 & 43.9 \\
\rowcolor{OurColor}
\textbf{+ \ours (Ours)} & & 47.4 & 12.2 & & \textbf{6.1} & \textbf{52.9} & 30.3 & \textbf{3.2} & & \textbf{85.7} & 91.7 & \textbf{86.2} & \textbf{88.9} & & \textbf{14.7} & \textbf{56.0} & \textbf{45.6} \\
\midrule
Qwen2.5-14B & & 42.6 & 12.2 & & 6.1 & \textbf{52.3} & 29.4 & 3.6 & & 86.2 & 87.9 & 91.8 & 89.8 & & 23.7 & 61.5 & \textbf{56.5} \\
\rowcolor{OurColor}
\textbf{+ \ours (Ours)} & & \textbf{39.0} & \textbf{10.6} & & \textbf{5.4} & 51.9 & \textbf{28.0} & \textbf{3.1} & & \textbf{89.0} & \textbf{90.4} & \textbf{93.4} & \textbf{91.9} & & \textbf{25.9} & \textbf{66.2} & 56.4 \\
\midrule
Qwen3-14B & & 43.2 & 11.7 & & 5.5 & \textbf{53.1} & 28.8 & 3.2 & & 88.8 & \textbf{91.6} & 91.5 & 91.5 & & 23.7 & 66.2 & 53.0 \\
\rowcolor{OurColor}
\textbf{+ \ours (Ours)} & & \textbf{41.2} & \textbf{11.4} & & \textbf{5.0} & 53.0 & \textbf{26.7} & \textbf{3.0} & & \textbf{88.9} & 90.7 & \textbf{92.7} & \textbf{91.7} & & \textbf{27.9} & \textbf{69.0} & \textbf{56.8} \\
\bottomrule
\end{tabular}
}
\vspace{-0.2cm}
\end{table}

\begin{table}[t]
\centering
\caption{VQA comparison of different representation alignment strategies for MLLMs.}
\label{tab:comparison}
\vspace{-0.1cm}
\setlength{\tabcolsep}{.42em}
\resizebox{0.92\linewidth}{!}{%
\begin{tabular}{l c c c c c c c >{\columncolor{OurColor!30}}c>{\columncolor{OurColor!30}}c>{\columncolor{OurColor!30}}c>{\columncolor{OurColor!30}}c>{\columncolor{OurColor!30}}c>{\columncolor{OurColor!30}}c>{\columncolor{OurColor!30}}c}
\toprule 
& & \textbf{General} & & \textbf{Knowledge} & & \textbf{OCR} & & \multicolumn{7}{c}{\cellcolor{OurColor!30}\textbf{Vision}} \\
\cmidrule(lr){3-3} \cmidrule(lr){5-5} \cmidrule(lr){7-7} \cmidrule{9-15}
\textbf{Alignment} & & Avg & & Avg & & Avg & & RWQA & MMVP & Blink & V* & CVBench & Avg & \\
\midrule
- & & 74.7 & & 49.7 & & 43.5 & & 59.2 & 49.3 & 52.2 & 50.8 & 67.8 & 55.9 & \\
ROSS~\cite{wangreconstructive} & & 74.6 & & 49.4 & & 44.0 & & 59.8 & 50.3 & 52.2 & 49.2 & 69.7 & 56.2 & \inc{0.3} \\
VIRAL~\cite{yoon2025visual} & & 73.8 & & 49.3 & & 43.6 & & 57.6 & 46.7 & 51.3 & 46.1 & 69.3 & 54.2 & \dec{1.7} \\
JARVIS~\cite{caffagni2025seeing} & & 76.8 & & 49.9 & & 46.2 & & 59.2 & 54.7 & 54.2 & \textbf{55.5} & 69.9 & 58.7 & \inc{2.8} \\
CMAR~\cite{gan2025cross} & & 76.4 & & 51.0 & & 46.0 & & 58.8 & 50.0 & 52.1 & 52.9 & 70.9 & 56.9 & \inc{1.0} \\
\rowcolor{OurColor}
\textbf{\ours (Ours)} & & \textbf{76.9} & & \textbf{51.1} & & \textbf{47.6} & & \textbf{60.1} & \textbf{58.0} & \textbf{55.5} & 50.3 & \textbf{73.5} & \textbf{59.5} & \inc{3.6} \\
\bottomrule
\end{tabular}
}
\vspace{-0.35cm}
\end{table}

\subsection{Main Experimental Results} 

\tit{Results on Cambrian Benchmarks}
To assess the generalizability and scalability of our proposed representation alignment regularization, we evaluate \ours across a diverse suite of language models, as reported in Table~\ref{tab:llms}. We deliberately select models spanning multiple architectural generations to ensure our findings are not isolated to a specific design. This includes established baselines like the Vicuna family, as well as the latest generation of state-of-the-art open-source models, such as Qwen3. Furthermore, we scale the parameter count across our experiments, progressing from compact models (3B and 4B) up to larger reasoning engines (13B and 14B). We remind to Appendix~\ref{sec:appendix_results} for the complete breakdown of the General, Knowledge, and OCR categories.

A common risk when forcefully modifying the internal representations of a pre-trained LLM is the potential degradation of its inherent linguistic and reasoning priors. However, the results demonstrate that our head-wise alignment successfully preserves, and frequently improves, the model's core competencies. Across the General, Knowledge, and OCR task categories, the inclusion of \ours consistently yields stable or higher average scores compared to the standard LLaVA training recipe. For instance, Qwen2.5-14B sees its General average rise from 75.6 to 77.4, while its Knowledge and OCR averages experience parallel uplifts. This indicates that isolating the alignment to a strategic subset of attention heads successfully mitigates catastrophic interference with the main language modeling objective.

The most substantial impact of \ours is observed in the Vision-Centric benchmarks, which directly measure visual perception, spatial reasoning, and multimodal grounding. Regardless of the underlying architecture or its release date, our method systematically drives up visual performance. Earlier models like Vicuna-7B experience a robust +2.3 point gain, proving that proper representation alignment can benefit legacy architectures. Simultaneously, modern models equipped with stronger text priors also reap significant benefits; notably, Qwen3-8B achieves the highest individual leap with a +3.6 average improvement. 

This trend persists as we scale the LLM backbone. When applied to the largest models in our suite, the representation alignment remains highly effective, with Qwen2.5-14B securing a $+3.4$ point increase and Qwen3-14B pushing the upper bound of the Vision-Centric average to 58.9.

\tit{Results on Hallucination Benchmarks} 
Although mitigating hallucinations, an open problem in MLLMs, is not explicitly enforced by our contrastive representation alignment loss, we find that \ours has a positive effect on it, as outlined in Table~\ref{tab:hall}. Across both the CHAIR-MSCOCO and AMBER generative benchmarks, models trained with \ours consistently lower their hallucination rates (\eg, CHAIR$_s$ and CHAIR$_i$). Crucially, on AMBER, this reduction in hallucinations is achieved while simultaneously improving, or at least maintaining, the cognition (Cog) score, which is a challenging balance, as models often become overly conservative when penalized for hallucinations. 

In discriminative settings, \ours yields steady improvements in accuracy and F1 scores on AMBER, indicating a more robust visual grounding. On HallusionBench, the method drives positive gains across nearly all models, significantly improving qAcc, Easy, and Hard metrics. The sole exception is Qwen3-4B, with a drop on this specific benchmark; however, this is vastly offset by the superior performance gains across standard VQA and Vision-Centric benchmarks (as detailed in Table~\ref{tab:llms}). Ultimately, explicitly aligning LLM internal representations with the visual domain naturally curbs the tendency to over-rely on linguistic priors, resulting in more faithful vision-language generations.

\tit{Comparison with Previous Representation Alignment Methods} 
In Table~\ref{tab:comparison}, we compare \ours against recent representation alignment strategies: ROSS~\cite{wangreconstructive} trains an auxiliary denoiser network conditioned on LLM visual features to recover visual tokens; VIRAL~\cite{yoon2025visual} aligns visual features from the LLM middle layer during the instruction tuning stage; JARVIS~\cite{caffagni2025seeing} reconstructs masked image latents using representations from one quarter of the LLM depth; and CMAR~\cite{gan2025cross} optimizes the CKA alignment metric between textual features from the penultimate LLM layer and the teacher encoder. We run each experiment according to its official implementation. All methods are trained on the same LLaVA~\cite{liu2024improved} dataset, feature Qwen3-8B~\cite{bai2025qwen3} as the LLM, SigLIP2 ViT-SO400M/14@384~\cite{tschannen2025siglip} as the vision encoder, and DINOv2-L~\cite{oquab2024dinov2} as the teacher for alignment (with the exception of ROSS).

Compared to these fixed-layer approaches, our targeted head-wise alignment proves significantly more effective. Notably, the strict pointwise feature alignment operated by VIRAL is the only method that registers a regression compared to LLaVA (\textit{first} row). Furthermore, while CMAR shares our goal of topological alignment between spaces of different modalities, the CKA metric forces \textit{global} point-wise relationships to match those from vision encoder. By contrast, \ours focuses strictly on preserving \textit{local} neighborhood structures, without imposing rigid distance constraints between samples. Ultimately, on the demanding Vision-Centric benchmarks, \ours yields a +3.6 point average improvement, outperforming the next best method, JARVIS (+2.8), and achieves the highest overall scores across the General, Knowledge, and OCR tasks. Full detailed results for each category are provided in Appendix~\ref{sec:appendix_results}.

\subsection{Varying the Teacher Visual Encoder}
\begin{figure}[t]
    \centering
    \includegraphics[width=\linewidth]{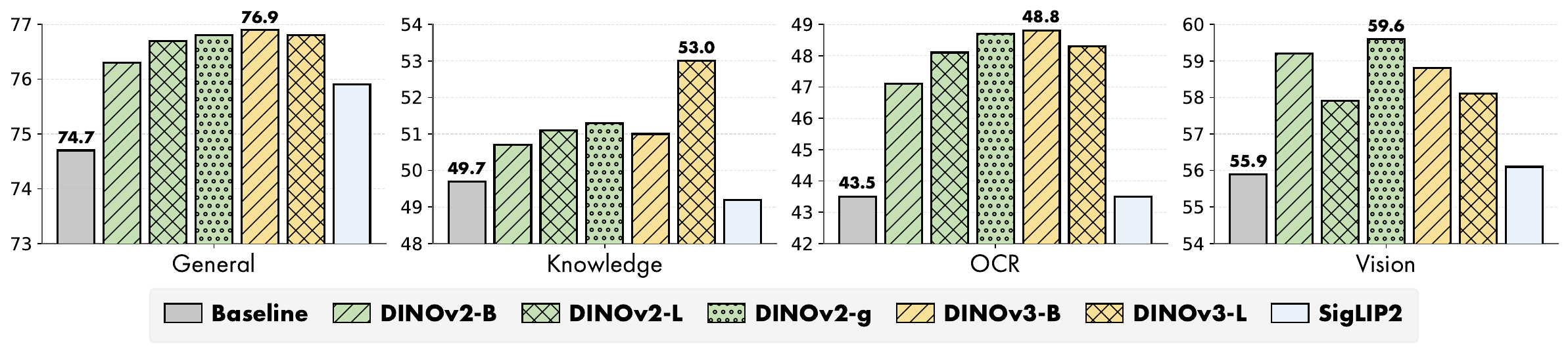}
    \vspace{-0.7cm}
    \caption{VQA results of \ours with different teacher vision encoders.}
    \label{fig:teacher}
    \vspace{-0.35cm}
\end{figure}

In Fig.~\ref{fig:teacher}, we experiment with different teacher vision encoders used to extract the targets for the \ours contrastive loss. All models are trained using Qwen3-8B as the language backbone and SigLIP2 as the primary vision encoder. We observe that using SigLIP2 itself as the teacher is mostly ineffective, in accordance with concurrent work~\cite{yoon2025visual,yurepresentation} showing that unsupervised vision encoders are better representation teachers than encoders trained with language supervision. Indeed, aligning with DINO-based~\cite{oquab2024dinov2,simeoni2025dinov3} teachers yields strong and consistent gains, even when using the base models (DINOv2-B and DINOv3-B). However, we note no clear benefits from employing the larger 1B-parameter DINOv2-g, suggesting that base vision encoders may suffice for topological alignment.

\section{Conclusion}
\label{sec:conclusion}
In this work, we introduced \textbf{\underline{He}}ad-Wise \textbf{\underline{R}}epresentation \textbf{\underline{A}}lignment (\textbf{\ours}), a novel method to enhance Multimodal Large Language Models through topological representation alignment. Guided by the Platonic Representation Hypothesis, \ours uses a contrastive proxy for the MKNN metric to align specific attention heads with an external vision encoder, demonstrating that targeting the \textit{least} aligned heads yields the most substantial gains. Evaluations across multiple architectures and benchmarks reveal that our approach significantly benefits demanding vision-centric tasks without compromising, even improving, core linguistic capabilities, while mitigating visual hallucinations.

\clearpage
\section*{Acknowledgments}
This work has been supported by the EU Horizon project ``ELLIOT'' (No. 101214398), by the EuroHPC JU project ``MINERVA'' (GA No. 101182737), and by the PNRR project ``ITSERR'' (CUP B53C22001770006) funded by the EU - NextGenerationEU. We also acknowledge EuroHPC JU for awarding the project EHPC-AIF-2025SC04-225 access to LUMI at CSC, Finland.

\bibliographystyle{plain}
\bibliography{bibliography}

@string{cvpr     = {CVPR}}

@string{nips     = {NeurIPS}}

@string{iccv     = {ICCV}}

@string{eccv     = {ECCV}}

@string{iclr     = {ICLR}}

@string{iclrw    = {ICLR Workshops}}

@string{icml     = {ICML}}

@string{emnlp    = {EMNLP}}

@string{acl      = {ACL}}

@string{tmlr      = {TMLR}}

@inproceedings{liu2024improved,
  title={{Improved Baselines with Visual Instruction Tuning}},
  author={Liu, Haotian and Li, Chunyuan and Li, Yuheng and Lee, Yong Jae},
  booktitle=cvpr,
  year={2024}
}

@inproceedings{huh2024platonic,
  title={{The Platonic Representation Hypothesis}},
  author={Huh, Minyoung and Cheung, Brian and Wang, Tongzhou and Isola, Phillip},
  booktitle=icml,
  year={2024}
}

@article{olsson2022context,
  title={{In-Context Learning and Induction Heads}},
  author={Olsson, Catherine and Elhage, Nelson and Nanda, Neel and Joseph, Nicholas and DasSarma, Nova and Henighan, Tom and Mann, Ben and Askell, Amanda and Bai, Yuntao and Chen, Anna and others},
  journal={arXiv preprint arXiv:2209.11895},
  year={2022}
}

@inproceedings{wanginterpretability,
  title={{Interpretability in the Wild: a Circuit for Indirect Object Identification in GPT-2 Small}},
  author={Wang, Kevin Ro and Variengien, Alexandre and Conmy, Arthur and Shlegeris, Buck and Steinhardt, Jacob},
  booktitle=iclr,
  year={2023}
}

@misc{vicuna2023,
    title = {{Vicuna: An Open-Source Chatbot Impressing GPT-4 with 90\%* ChatGPT Quality}},
    author = {Chiang, Wei-Lin and Li, Zhuohan and Lin, Zi and Sheng, Ying and Wu, Zhanghao and Zhang, Hao and Zheng, Lianmin and Zhuang, Siyuan and Zhuang, Yonghao and Gonzalez, Joseph E. and Stoica, Ion and Xing, Eric P.},
    year = {2023}
}

@article{grattafiori2024llama,
  title={{The Llama 3 Herd of Models}},
  author={Grattafiori, Aaron and Dubey, Abhimanyu and Jauhri, Abhinav and Pandey, Abhinav and Kadian, Abhishek and Al-Dahle, Ahmad and Letman, Aiesha and Mathur, Akhil and Schelten, Alan and Vaughan, Alex and others},
  journal={arXiv preprint arXiv:2407.21783},
  year={2024}
}

@inproceedings{yue2024less,
  title={{Less is More: Mitigating Multimodal Hallucination from an EOS Decision Perspective}},
  author={Yue, Zihao and Zhang, Liang and Jin, Qin},
  booktitle=acl,
  year={2024}
}

@article{wang2023amber,
  title={{AMBER: An LLM-free Multi-dimensional Benchmark for MLLMs Hallucination Evaluation}},
  author={Wang, Junyang and Wang, Yuhang and Xu, Guohai and Zhang, Jing and Gu, Yukai and Jia, Haitao and Wang, Jiaqi and Xu, Haiyang and Yan, Ming and Zhang, Ji and others},
  journal={arXiv preprint arXiv:2311.07397},
  year={2023}
}

@inproceedings{namcausal,
  title={{Causal Head Gating: A Framework for Interpreting Roles of Attention Heads in Transformers}},
  author={Nam, Andrew Joohun and Conklin, Henry and Yang, Yukang and Griffiths, Thomas L and Cohen, Jonathan D and Leslie, Sarah-Jane},
  booktitle=nips,
  year={2025}
}

@inproceedings{yurepresentation,
  title={{Representation Alignment for Generation: Training Diffusion Transformers Is Easier Than You Think}},
  author={Yu, Sihyun and Kwak, Sangkyung and Jang, Huiwon and Jeong, Jongheon and Huang, Jonathan and Shin, Jinwoo and Xie, Saining},
  booktitle=iclr,
  year=2025
}

@article{yoon2025visual,
  title={{Visual Representation Alignment for Multimodal Large Language Models}},
  author={Yoon, Heeji and Jung, Jaewoo and Kim, Junwan and Choi, Hyungyu and Shin, Heeseong and Lim, Sangbeom and An, Honggyu and Kim, Chaehyun and Han, Jisang and Kim, Donghyun and others},
  journal={arXiv preprint arXiv:2509.07979},
  year={2025}
}

@inproceedings{leng2025repa,
  title={{REPA-E: Unlocking VAE for End-to-End Tuning with Latent Diffusion Transformers}},
  author={Leng, Xingjian and Singh, Jaskirat and Hou, Yunzhong and Xing, Zhenchang and Xie, Saining and Zheng, Liang},
  booktitle=iccv,
  year=2025
}

@inproceedings{groger2026revisiting,
  title={{Revisiting the Platonic Representation Hypothesis: An Aristotelian View}},
  author={Gr{\"o}ger, Fabian and Wen, Shuo and Brbi{\'c}, Maria},
  booktitle=icml,
  year=2026
}

@article{bai2025qwen3,
  title={{Qwen3-VL Technical Report}},
  author={Bai, Shuai and Cai, Yuxuan and Chen, Ruizhe and Chen, Keqin and Chen, Xionghui and Cheng, Zesen and Deng, Lianghao and Ding, Wei and Gao, Chang and Ge, Chunjiang and others},
  journal={arXiv preprint arXiv:2511.21631},
  year={2025}
}

@article{wang2025internvl3,
  title={{InternVL3.5: Advancing Open-Source Multimodal Models in Versatility, Reasoning, and Efficiency}},
  author={Wang, Weiyun and Gao, Zhangwei and Gu, Lixin and Pu, Hengjun and Cui, Long and Wei, Xingguang and Liu, Zhaoyang and Jing, Linglin and Ye, Shenglong and Shao, Jie and others},
  journal={arXiv preprint arXiv:2508.18265},
  year={2025}
}

@inproceedings{tong2024cambrian,
  title={{Cambrian-1: A Fully Open, Vision-Centric Exploration of Multimodal LLMs}},
  author={Tong, Shengbang and Brown, Ellis and Wu, Penghao and Woo, Sanghyun and Middepogu, Manoj and Akula, Sai C and Yang, Jihan and Yang, Shusheng and Iyer, Adithya and Pan, Xichen and others},
  booktitle=nips,
  year={2024}
}

@article{caffagni2025seeing,
  title={{Seeing Beyond Words: Self-Supervised Visual Learning for Multimodal Large Language Models}},
  author={Caffagni, Davide and Sarto, Sara and Cornia, Marcella and Baraldi, Lorenzo and Dovesi, Pier Luigi and Roohi, Shaghayegh and Granroth-Wilding, Mark and Cucchiara, Rita},
  journal={arXiv preprint arXiv:2512.15885},
  year={2025}
}

@inproceedings{gan2025cross,
  title={{Cross-Modal Alignment Regularization: Enhancing Language Models with Vision Model Representations}},
  author={Gan, Yulu and Zhao, Kaiya Ivy and Isola, Phillip},
  booktitle=iclrw,
  year={2025}
}

@inproceedings{wangreconstructive,
  title={{Reconstructive Visual Instruction Tuning}},
  author={Wang, Haochen and Zheng, Anlin and Zhao, Yucheng and Wang, Tiancai and Ge, Zheng and Zhang, Xiangyu and Zhang, Zhaoxiang},
  booktitle=iclr,
  year={2025}
}

@article{qwen2024qwen25technicalreport,
      title={{Qwen2.5 Technical Report}}, 
      author={{Qwen Team}},
      journal={arXiv preprint arXiv:2412.15115},
      year={2024}
}

@article{yang2025qwen3,
  title={{Qwen3 Technical Report}},
  author={Yang, An and Li, Anfeng and Yang, Baosong and Zhang, Beichen and Hui, Binyuan and Zheng, Bo and Yu, Bowen and Gao, Chang and Huang, Chengen and Lv, Chenxu and others},
  journal={arXiv preprint arXiv:2505.09388},
  year={2025}
}

@inproceedings{guan2024hallusionbench,
  title={{HallusionBench: An Advanced Diagnostic Suite for Entangled Language Hallucination and Visual Illusion in Large Vision-Language Models}},
  author={Guan, Tianrui and Liu, Fuxiao and Wu, Xiyang and Xian, Ruiqi and Li, Zongxia and Liu, Xiaoyu and Wang, Xijun and Chen, Lichang and Huang, Furong and Yacoob, Yaser and others},
  booktitle=cvpr,
  year={2024}
}

@inproceedings{hanoona2023GLaMM,
title={{GLaMM: Pixel Grounding Large Multimodal Model}},
author={Rasheed, Hanoona and Maaz, Muhammad and Shaji, Sahal and Shaker, Abdelrahman and Khan, Salman and Cholakkal, Hisham and Anwer, Rao M. and Xing, Eric and Yang, Ming-Hsuan and Khan, Fahad S.},
booktitle=cvpr,
year={2024}
}

@article{tschannen2025siglip,
  title={{SigLIP 2: Multilingual Vision-Language Encoders with Improved Semantic Understanding, Localization, and Dense Features}},
  author={Tschannen, Michael and Gritsenko, Alexey and Wang, Xiao and Naeem, Muhammad Ferjad and Alabdulmohsin, Ibrahim and Parthasarathy, Nikhil and Evans, Talfan and Beyer, Lucas and Xia, Ye and Mustafa, Basil and others},
  journal={arXiv preprint arXiv:2502.14786},
  year={2025}
}

@article{oquab2024dinov2,
  title={{DINOv2: Learning Robust Visual Features without Supervision}},
  author={Oquab, Maxime and Darcet, Timoth{\'e}e and Moutakanni, Th{\'e}o and Vo, Huy and Szafraniec, Marc and Khalidov, Vasil and Fernandez, Pierre and Haziza, Daniel and Massa, Francisco and El-Nouby, Alaaeldin and others},
  journal=tmlr,
  pages={1--31},
  year={2024}
}

@misc{grok,
  title={Grok},
  author={xAI},
  year={2024},
  url={https://x.ai/blog/grok-1.5v}
}

@inproceedings{fu2024blink,
  title={{BLINK: Multimodal Large Language Models Can See But Not Perceive}},
  author={Fu, Xingyu and Hu, Yushi and Li, Bangzheng and Feng, Yu and Wang, Haoyu and Lin, Xudong and Roth, Dan and Smith, Noah A and Ma, Wei-Chiu and Krishna, Ranjay},
  booktitle=eccv,
  year={2024},
}

@inproceedings{tong2024eyes,
  title={{Eyes Wide Shut? Exploring the Visual Shortcomings of Multimodal LLMs}},
  author={Tong, Shengbang and Liu, Zhuang and Zhai, Yuexiang and Ma, Yi and LeCun, Yann and Xie, Saining},
  booktitle=cvpr,
  year={2024}
}

@article{fu2023mme,
  title={{MME: A Comprehensive Evaluation Benchmark for Multimodal Large Language Models}},
  author={Fu, Chaoyou and Chen, Peixian and Shen, Yunhang and Qin, Yulei and Zhang, Mengdan and Lin, Xu and Yang, Jinrui and Zheng, Xiawu and Li, Ke and Sun, Xing and others},
  journal={arXiv preprint arXiv:2306.13394},
  year={2023}
}

@article{li2023seed,
  title={{SEED-Bench: Benchmarking Multimodal LLMs with Generative Comprehension}},
  author={Li, Bohao and Wang, Rui and Wang, Guangzhi and Ge, Yuying and Ge, Yixiao and Shan, Ying},
  journal={arXiv preprint arXiv:2307.16125},
  year={2023}
}

@inproceedings{hudson2019gqa,
  title={{GQA: A New Dataset for Real-World Visual Reasoning and Compositional Question Answering}},
  author={Hudson, Drew A and Manning, Christopher D},
  booktitle=cvpr,
  year={2019}
}

@inproceedings{lu2022learn,
  title={{Learn to Explain: Multimodal Reasoning via Thought Chains for Science Question Answering}},
  author={Lu, Pan and Mishra, Swaroop and Xia, Tanglin and Qiu, Liang and Chang, Kai-Wei and Zhu, Song-Chun and Tafjord, Oyvind and Clark, Peter and Kalyan, Ashwin},
  booktitle=nips,
  year={2022}
}

@inproceedings{lu2023mathvista,
  title={{MathVista: Evaluating Mathematical Reasoning of Foundation Models in Visual Contexts}},
  author={Lu, Pan and Bansal, Hritik and Xia, Tony and Liu, Jiacheng and Li, Chunyuan and Hajishirzi, Hannaneh and Cheng, Hao and Chang, Kai-Wei and Galley, Michel and Gao, Jianfeng},
  booktitle=iclr,
  year={2024}
}

@inproceedings{yue2023mmmu,
  title={{MMMU: A Massive Multi-discipline Multimodal Understanding and Reasoning Benchmark for Expert AGI}},
  author={Yue, Xiang and Ni, Yuansheng and Zhang, Kai and Zheng, Tianyu and Liu, Ruoqi and Zhang, Ge and Stevens, Samuel and Jiang, Dongfu and Ren, Weiming and Sun, Yuxuan and others},
  booktitle=cvpr,
  year={2024}
}

@inproceedings{masry2022chartqa,
  title={{ChartQA: A Benchmark for Question Answering about Charts with Visual and Logical Reasoning}},
  author={Masry, Ahmed and Do, Xuan Long and Tan, Jia Qing and Joty, Shafiq and Hoque, Enamul},
  booktitle=acl,
  year={2022}
}

@article{liu2024ocrbench,
  title={{OCRBench: On the Hidden Mystery of OCR in Large Multimodal Models}},
  author={Liu, Yuliang and Li, Zhang and Huang, Mingxin and Yang, Biao and Yu, Wenwen and Li, Chunyuan and Yin, Xu-Cheng and Liu, Cheng-Lin and Jin, Lianwen and Bai, Xiang},
  journal={Sci China Inf Sci},
  volume={67},
  number={12},
  pages={220102},
  year={2024},
}

@inproceedings{liu2024mmbench,
  title={{MMBench: Is Your Multi-modal Model an All-around Player?}},
  author={Liu, Yuan and Duan, Haodong and Zhang, Yuanhan and Li, Bo and Zhang, Songyang and Zhao, Wangbo and Yuan, Yike and Wang, Jiaqi and He, Conghui and Liu, Ziwei and others},
  booktitle=eccv,
  year={2024},
}

@inproceedings{kembhavi2016diagram,
  title={{A diagram is worth a dozen images}},
  author={Kembhavi, Aniruddha and Salvato, Mike and Kolve, Eric and Seo, Minjoon and Hajishirzi, Hannaneh and Farhadi, Ali},
  booktitle=eccv,
  year={2016},
}

@inproceedings{lin2014microsoft,
  title={{Microsoft COCO: Common Objects in Context}},
  author={Lin, Tsung-Yi and Maire, Michael and Belongie, Serge and Hays, James and Perona, Pietro and Ramanan, Deva and Doll{\'a}r, Piotr and Zitnick, C Lawrence},
  booktitle=eccv,
  year={2014},
}

@inproceedings{wu2024v,
  title={{V*: Guided Visual Search as a Core Mechanism in Multimodal LLMs}},
  author={Wu, Penghao and Xie, Saining},
  booktitle=cvpr,
  year={2024}
}

@inproceedings{li2023evaluating,
  title={{Evaluating Object Hallucination in Large Vision-Language Models}},
  author={Li, Yifan and Du, Yifan and Zhou, Kun and Wang, Jinpeng and Zhao, Xin and Wen, Ji-Rong},
  booktitle=emnlp,
  year={2023}
}

@inproceedings{singh2019towards,
  title={{Towards VQA Models That Can Read}},
  author={Singh, Amanpreet and Natarajan, Vivek and Shah, Meet and Jiang, Yu and Chen, Xinlei and Batra, Dhruv and Parikh, Devi and Rohrbach, Marcus},
  booktitle=cvpr,
  year={2019}
}

@misc{qwen35towards,
    title = {{Qwen3.5: Towards Native Multimodal Agents}},
    url = {https://qwen.ai/blog?id=qwen3.5},
    author = {{Qwen Team}},
    year = {2026}
}

@article{tong2026beyond,
  title={{Beyond Language Modeling: An Exploration of Multimodal Pretraining}},
  author={Tong, Shengbang and Fan, David and Nguyen, John and Brown, Ellis and Zhou, Gaoyue and Qian, Shengyi and Zheng, Boyang and Vallaeys, Th{\'e}ophane and Han, Junlin and Fergus, Rob and others},
  journal={arXiv preprint arXiv:2603.03276},
  year={2026}
}

@inproceedings{gurari2018vizwiz,
  title={{VizWiz Grand Challenge: Answering Visual Questions From Blind People}},
  author={Gurari, Danna and Li, Qing and Stangl, Abigale J and Guo, Anhong and Lin, Chi and Grauman, Kristen and Luo, Jiebo and Bigham, Jeffrey P},
  booktitle=cvpr,
  year={2018}
}

@article{simeoni2025dinov3,
  title={{DINOv3}},
  author={Sim{\'e}oni, Oriane and Vo, Huy V and Seitzer, Maximilian and Baldassarre, Federico and Oquab, Maxime and Jose, Cijo and Khalidov, Vasil and Szafraniec, Marc and Yi, Seungeun and Ramamonjisoa, Micha{\"e}l and others},
  journal={arXiv preprint arXiv:2508.10104},
  year={2025}
}

@article{oord2018representation,
  title={{Representation Learning with
Contrastive Predictive Coding}},
  author={Oord, Aaron van den and Li, Yazhe and Vinyals, Oriol},
  journal={arXiv preprint arXiv:1807.03748},
  year={2018}
}

@inproceedings{fan2025scaling,
  title={{Scaling Language-Free Visual Representation Learning}},
  author={Fan, David and Tong, Shengbang and Zhu, Jiachen and Sinha, Koustuv and Liu, Zhuang and Chen, Xinlei and Rabbat, Michael and Ballas, Nicolas and LeCun, Yann and Bar, Amir and others},
  booktitle=iccv,
  year={2025}
}

@article{wiedmann2025finevision,
  title={{FineVision: Open Data Is All You Need}},
  author={Wiedmann, Luis and Zohar, Orr and Mahla, Amir and Wang, Xiaohan and Li, Rui and Frere, Thibaud and von Werra, Leandro and Gosthipaty, Aritra Roy and Marafioti, Andr{\'e}s},
  journal={arXiv preprint arXiv:2510.17269},
  year={2025}
}


\clearpage
\appendix

\section{Additional Implementation Details}
\label{sec:appendix_details}

\tit{Training Details} Following the two-stage training recipe of LLaVA-1.5~\cite{liu2024improved}, in the first stage, we train only the projector $\proj$, a two-layer MLP, using 558k image-caption pairs, while keeping the language model $\llm$ frozen. In the second stage, we jointly optimize $\llm$ and $\proj$ for visual instruction tuning on the LLaVA-Instruct-665k dataset. The training settings (\eg, optimizer, learning rate, and batch size) are kept identical to LLaVA-1.5. All experiments are conducted on AMD MI250x devices, each of which comprises 2 GPUs with 64GB of VRAM. The first training stage runs on 16 GPUs for up to 6 hours, depending on the size of the LLM. The second training stage runs on 32 GPUs, up to 14 hours. We find no noticeable difference in training time with the addition of $\losshera$. 
The MKNN alignment scores can be efficiently computed offline. For instance, for a 7B LLM and DINOv2-L, it takes less than one hour on a single GPU.

\tit{Contrastive Learning Details}
To generate supervision signals for each batch, we extract the \texttt{[CLS]} token representations from the teacher vision encoder $\teachvisenc$. We then compute the pairwise dot products between these representations to form a similarity matrix (\ie, a Gram matrix), which captures the visual neighborhood structure of the batch. For every sample, we identify its top-$k$ nearest neighbors to construct multi-positive contrastive targets. This is achieved by assigning a uniform probability of $\frac{1}{k}$ to these $k$ neighbors, and a probability of zero to all other samples.

For the student model, we extract the head-wise representations from the selected set of heads ($\heads$) and apply average pooling across the embeddings corresponding to text tokens. In the first training stage, text tokens correspond to the caption of the input image, and thus we employ all of them. In the second training stage, text tokens represent multi-turn dialogs, and we pool exclusively over the tokens pertaining to the \texttt{<ASSISTANT>} turn. These are the same tokens contributing to the language modeling loss of Eq.~\ref{eq:loss_lm}. 

The temperature parameter $\tau$ of Eq.~\ref{eq:loss_ra} is learned in logarithmic scale, and it is initialized as 0.07.

\tit{LLM Details and Selected Heads} We collect in Table~\ref{tab:llms_details} the exact checkpoints of each LLM used in this work. All of them are publicly accessible on the Hugging Face Hub. We also report the specific attention heads of each LLM used to compute $\losshera$, sorted from left to right by increasing value of the MKNN alignment score. We indicate with \texttt{LXHY} the index of the \texttt{Y}-th head in layer \texttt{X}.

 \begin{table}[h]
\caption{Checkpoint reference and list of selected heads for each LLM.}
\label{tab:llms_details}
\centering
\vspace{-0.1cm}
\setlength{\tabcolsep}{0.4em}
\resizebox{0.98\linewidth}{!}{
\begin{tabular}{llccccc}
\toprule
 &  & \multicolumn{5}{c}{\textbf{Selected Heads ($\heads$})}\\
\cmidrule(lr){3-7}
\textbf{LLM} & \textbf{Hugging Face Page} & H1 & H2 & H3 & H4 & H5 \\
\midrule
Qwen2.5-3B~\cite{qwen2024qwen25technicalreport} &
{\small\texttt{\href{https://huggingface.co/Qwen/Qwen2.5-3B-Instruct}{Qwen/Qwen2.5-3B-Instruct}}} 
& L3H4 & L5H11 & L5H14 & L5H9 & L5H15 \\

Qwen3-4B~\cite{yang2025qwen3} &
{\small\texttt{\href{https://huggingface.co/Qwen/Qwen3-4B-Instruct-2507}{Qwen/Qwen3-4B-Instruct-2507}}} 
& L7H6 & L13H12 & L3H6 & L3H5 & L13H13 \\

Vicuna-7B~\cite{vicuna2023} &
{\small\texttt{\href{https://huggingface.co/lmsys/vicuna-7b-v1.5}{lmsys/vicuna-7b-v1.5}}} 
& L0H14 & L1H1 & L0H1 & L0H3 & L0H30 \\

Llama3-8B~\cite{grattafiori2024llama} &
{\small\texttt{\href{https://huggingface.co/meta-llama/Meta-Llama-3-8B-Instruct}{meta-llama/Meta-Llama-3-8B-Instruct}}} 
& L0H31 & L2H21 & L0H29 & L2H23 & L2H25 \\

Qwen2.5-7B~\cite{qwen2024qwen25technicalreport} &
{\small\texttt{\href{https://huggingface.co/Qwen/Qwen2.5-7B-Instruct}{Qwen/Qwen2.5-7B-Instruct}}} 
& L14H5 & L4H7 & L14H4 & L4H10 & L4H13 \\

Qwen3-8B~\cite{yang2025qwen3} &
{\small\texttt{\href{https://huggingface.co/Qwen/Qwen3-8B}{Qwen/Qwen3-8B}}} 
& L13H12 & L7H6 & L3H5 & L13H13 & L13H15 \\

Vicuna-13B~\cite{vicuna2023} &
{\small\texttt{\href{https://huggingface.co/lmsys/vicuna-13b-v1.5}{lmsys/vicuna-13b-v1.5}}} 
& L0H38 & L1H0 & L0H11 & L0H10 & L0H26 \\

Qwen2.5-14B~\cite{qwen2024qwen25technicalreport} &
{\small\texttt{\href{https://huggingface.co/Qwen/Qwen2.5-14B-Instruct}{Qwen/Qwen2.5-14B-Instruct}}} 
& L13H27 & L18H13 & L0H9 & L0H7 & L3H8 \\

Qwen3-14B~\cite{yang2025qwen3} &
{\small\texttt{\href{https://huggingface.co/Qwen/Qwen3-14B}{Qwen/Qwen3-14B}}} 
& L14H24 & L9H36 & L9H32 & L1H37 & L1H36 \\

\bottomrule
\end{tabular}
}
\end{table}

\section{Evaluation Benchmarks}
\label{sec:appendix_benchmarks}

\tit{Cambrian Evaluation Suite~\cite{tong2024cambrian}}\footnote{\small\texttt{\href{https://github.com/cambrian-mllm/cambrian}{https://github.com/cambrian-mllm/cambrian}}} It comprises a comprehensive suite of benchmarks designed to evaluate diverse capabilities of MLLMs, including general perception, knowledge reasoning, OCR and chart understanding, and core visual abilities. Accordingly, the benchmarks are grouped into four categories: General, Knowledge, OCR, and Vision. In our experiments, we consider 18 benchmarks: GQA~\cite{hudson2019gqa}, POPE~\cite{li2023evaluating}, MME~\cite{fu2023mme}, MMBench (MMB)~\cite{liu2024mmbench}, and SEED-Bench (SEED)~\cite{li2023seed} for the General category; ScienceQA (SQA)~\cite{lu2022learn}, MMMU~\cite{yue2023mmmu}, MathVista~\cite{lu2023mathvista}, and AI2D~\cite{kembhavi2016diagram} for Knowledge; ChartQA~\cite{masry2022chartqa}, OCRBench (OCRB)~\cite{liu2024ocrbench}, TextVQA~\cite{singh2019towards}, and VizWiz~\cite{gurari2018vizwiz} for OCR; and MMVP~\cite{tong2024eyes}, RealWorldQA (RWQA)~\cite{grok}, Blink~\cite{fu2024blink}, V*~\cite{wu2024v}, and CVBench~\cite{tong2024cambrian} for Vision. When reporting averages, we normalize the MME score by dividing it by 20 to ensure consistency with the scale of the other benchmarks.

\tit{Hallucination Datasets} 
We evaluate hallucinatory tendencies on three widely used benchmarks: AMBER~\cite{wang2023amber}, CHAIR-MSCOCO~\cite{yue2024less}, and HallusionBench~\cite{guan2024hallusionbench}. CHAIR-MSCOCO measures object- and sentence-level hallucination rates (\ie, CHAIR$_i$ and CHAIR$_s$) on model-generated descriptions for 500 images sampled from the MSCOCO~\cite{lin2014microsoft} validation set. The AMBER generative task further introduces \textit{cognition} (Cog), which quantifies the overlap between model- and human-hallucinated objects, and \textit{coverage} (Cover), which measures object-level recall. Complementarily, the AMBER discriminative task captures a broader set of hallucination types, including attribute and relation hallucinations in addition to object existence, using a ground truth set of 1,004 manually annotated images. For CHAIR-MSCOCO and the AMBER generative task, we use a maximum generation length of 512 tokens with greedy decoding. For the AMBER discriminative task, we append the instruction \textit{``Answer only with Yes or No. Use exactly one word. Do not use commas, periods, or symbols.''} to each query to enforce binary (Yes/No) responses. We evaluate hallucination robustness on HallusionBench~\cite{guan2024hallusionbench} using an exact-match protocol. Specifically, we force the model to output unambiguous responses by appending the following instruction to each query: \textit{``Answer the question using a single word or phrase: Yes or No.''}
We report three standard metrics. \emph{qAcc} (Question Pair Accuracy) measures group-level consistency. A prediction is counted as correct under qAcc only if the model answers all questions within the same group correctly.
In addition, we report \emph{Easy} accuracy, computed over unmodified questions, and \emph{Hard} accuracy, computed over adversarially modified or misleading variants designed to induce hallucinations.

\begin{table}[t]
\centering
\caption{VQA comparison between DINOv2 and SigLIP2 as the vision encoder for LLaVA.}
\label{tab:ablation_supp}
\vspace{-0.1cm}
\setlength{\tabcolsep}{.35em}
\resizebox{\linewidth}{!}{%
\begin{tabular}{l c c c c c c c c >{\columncolor{OurColor!30}}c>{\columncolor{OurColor!30}}c>{\columncolor{OurColor!30}}c>{\columncolor{OurColor!30}}c>{\columncolor{OurColor!30}}c>{\columncolor{OurColor!30}}c}
\toprule 
& & & \textbf{General} & & \textbf{Knowledge} & & \textbf{OCR} & & \multicolumn{6}{c}{\cellcolor{OurColor!30}\textbf{Vision}} \\
\cmidrule(lr){4-4} \cmidrule(lr){6-6} \cmidrule(lr){8-8} \cmidrule{10-15}
\textbf{LLM} & \textbf{Vision Encoder} & & Avg & & Avg & & Avg & & RWQA & MMVP & Blink & V* & CVBench & Avg \\
\midrule
Qwen2.5-3B & DINOv2 & & 67.1 & & 43.3 & & 26.3 & & 51.5 & 30.0 & 47.8 & 46.1 & 58.0 & 46.7 \\
Qwen2.5-3B & SigLIP2 & & 73.5 & & 46.7 & & 42.2 & & 55.2 & 46.0 & 46.8 & 44.5 & \textbf{60.2} & 50.5 \\
\rowcolor{OurColor}
\textbf{+ \ours (Ours)} & SigLIP2 & & \textbf{74.5} & & \textbf{47.5} & & \textbf{43.8} & & \textbf{56.3} & \textbf{48.0} & \textbf{49.1} & \textbf{51.3} & 59.6 & \textbf{52.9} \\
\midrule
Qwen3-4B & DINOv2 & & 70.5 & & 46.1 & & 24.2 & & 54.9 & 38.7 & 49.2 & 49.2 & 63.1 & 51.0 \\
Qwen3-4B & SigLIP2 & & 75.6 & & 49.6 & & 43.8 & & 59.9 & 48.0 & \textbf{55.1} & 50.3 & 68.3 & 56.3 \\
\rowcolor{OurColor}
\textbf{+ \ours (Ours)} & SigLIP2 & & \textbf{76.0} & & \textbf{50.1} & & \textbf{44.5} & & \textbf{61.3} & \textbf{56.0} & 53.7 & \textbf{52.4} & \textbf{69.1} & \textbf{58.5} \\
\bottomrule
\end{tabular}
}
\vspace{-0.2cm}
\end{table}

\begin{table}[t]
\centering
\caption{Effects of $\losshera$ when applied to the different training stages of LLaVA.}
\label{tab:ablation_stages}
\vspace{-0.1cm}
\setlength{\tabcolsep}{.42em}
\resizebox{\linewidth}{!}{%
\begin{tabular}{l c c c c c c c c c >{\columncolor{OurColor!30}}c>{\columncolor{OurColor!30}}c>{\columncolor{OurColor!30}}c>{\columncolor{OurColor!30}}c>{\columncolor{OurColor!30}}c>{\columncolor{OurColor!30}}c}
\toprule 
& \multicolumn{2}{c}{$\losshera$} & & \textbf{General} & & \textbf{Knowledge} & & \textbf{OCR} & & \multicolumn{6}{c}{\cellcolor{OurColor!30}\textbf{Vision}} \\
\cmidrule(lr){2-3} \cmidrule(lr){5-5} \cmidrule(lr){7-7} \cmidrule(lr){9-9} \cmidrule{11-16}
\textbf{Model} & St.1 & St.2 & & Avg & & Avg & & Avg & & RWQA & MMVP & Blink & V* & CVBench & Avg \\
\midrule
Qwen2.5-3B & - & - & & 73.5 & & 46.7 & & 42.2 & & 55.2 & 46.0 & 46.8 & 44.5 & 60.2 & 50.5 \\
\textbf{+ \ours} & \checkmark & - & & 74.2 & & \textbf{47.5} & & \textbf{44.2} & & \textbf{57.4} & 44.0 & 46.9 & \textbf{51.3} & \textbf{60.7} & 52.1 \\
\textbf{+ \ours} & - & \checkmark & & 73.5 & & 46.4 & & 43.0 & & 54.5 & 40.0 & 48.8 & 44.5 & 60.5 & 49.7 \\
\rowcolor{OurColor}
\textbf{+ \ours (Ours)} & \checkmark & \checkmark & & \textbf{74.5} & & \textbf{47.5} & & 43.8 & & 56.3 & \textbf{48.0} & \textbf{49.1} & \textbf{51.3} & 59.6 & \textbf{52.9} \\
\midrule
Qwen3-4B & - & - & & 75.6 & & 49.6 & & 43.8 & & 59.9 & 48.0 & \textbf{55.1} & 50.3 & 68.3 & 56.3 \\
\textbf{+ \ours} & \checkmark & - & & 75.6 & & 49.7 & & 43.4 & & 60.7 & 51.3 & 53.3 & 46.1 & 66.2 & 55.5 \\
\textbf{+ \ours} & - & \checkmark & & 75.7 & & 50.0 & & \textbf{44.6} & & 59.6 & 54.7 & 52.7 & 49.2 & 67.8 & 56.8 \\
\rowcolor{OurColor}
\textbf{+ \ours (Ours)} & \checkmark & \checkmark & & \textbf{76.0} & & \textbf{50.1} & & 44.5 & & \textbf{61.3} & \textbf{56.0} & 53.7 & \textbf{52.4} & \textbf{69.1} & \textbf{58.5} \\
\bottomrule
\end{tabular}
}
\vspace{-0.3cm}
\end{table}

\section{Additional Experiments}
\label{sec:appendix_results}

\subsection{Additional Ablation Studies and Results}
  \tit{DINOv2 as Vision Encoder} In this work, we demonstrate the effectiveness of leveraging a teacher vision encoder, \eg, DINOv2-L~\cite{oquab2024dinov2}, as a source of supervision for topological representation alignment. It is natural to ask what if we do not perform representation alignment at all, by directly plugging in DINOv2-L as the vision encoder of an MLLM. Table~\ref{tab:ablation_supp} answers this question by comparing DINOv2-L vs SigLIP2~\cite{tschannen2025siglip}, and clearly demonstrates that DINOv2-L is ineffective on its own as a vision encoder for MLLMs. For fair comparison, we feed DINOv2-L with the same image resolution of $384 \times 384$ pixels as SigLIP2. Despite that, DINOv2-L suffers from severe deficits, especially on OCR tasks. These results agree with prior works~\cite{fan2025scaling,tong2024cambrian} showing that unsupervised visual encoders alone fall short against language-supervised encoders on VQA benchmarks. On the other hand, as testified by Fig.\ref{fig:teacher}, language-supervised encoders are unsuitable as representation teachers, and that justifies the application of representation alignment methods such as \ours, where MLLMs benefit from the synergistic effects of language-supervised and unsupervised visual encoders.

\tit{\ours in Different Training Stages} We seamlessly apply the $\losshera$ on both training stage of LLaVA. However, there are neat differences between them that are worth discussing. For instance, in the first training stage (\ie, St.1), the MLLM is fed with images and their related captions, which represent aligned image-text pairs, \ie, the same concept is expressed in two different modalities. This appears to be a suitable stage for representation alignment, as the student MLLM and the teacher vision encoder process the same underlying concepts. Conversely, image-text pairs in the second training stage (\ie, St.2) are not aligned the same way: the text corresponds to a multi-turn dialog between a user and the assistant, which focuses on the image, but does not exactly mimic the visual content as an image caption. With that in mind, if one had to select a single training stage for $\losshera$, we would expect a larger impact on St.1 rather than St.2. However, according to Table~\ref{tab:ablation_stages}, that is not always the case: with Qwen2.5-3B, $\losshera$ helps more when applied during St.1, while the opposite holds with Qwen3-4B. Ultimately, our original proposal of regularizing both training stages with $\losshera$ works best on both models.

\begin{table}[t]
\caption{Detailed VQA results of \ours applied to the LLaVA training recipe on different LLMs.}
\label{tab:llms_full}
\vspace{-0.1cm}
\centering
\setlength{\tabcolsep}{0.25em}
\resizebox{\linewidth}{!}{
\begin{tabular}{lc cccccc c ccccc c ccccc c >{\columncolor{OurColor!30}}c>{\columncolor{OurColor!30}}c>{\columncolor{OurColor!30}}c>{\columncolor{OurColor!30}}c>{\columncolor{OurColor!30}}c>{\columncolor{OurColor!30}}c}
\toprule
& & \multicolumn{6}{c}{\textbf{General}}
& & \multicolumn{5}{c}{\textbf{Knowledge}}
& & \multicolumn{5}{c}{\textbf{OCR}}
& & \multicolumn{6}{c}{\cellcolor{OurColor!30}\textbf{Vision}} \\
\cmidrule{3-8}
\cmidrule{10-14}
\cmidrule{16-20}
\cmidrule{22-27}
& & \rotatebox{65}{\textbf{Avg}} & \rotatebox{65}{GQA} & \rotatebox{65}{POPE} & \rotatebox{65}{MME} & \rotatebox{65}{MMB} & \rotatebox{65}{SEED} & & \rotatebox{65}{\textbf{Avg}} & \rotatebox{65}{SQA} & \rotatebox{65}{MMMU} & \rotatebox{65}{MathV} & \rotatebox{65}{AI2D} & & \rotatebox{65}{\textbf{Avg}} & \rotatebox{65}{ChartQA} & \rotatebox{65}{OCRB} & \rotatebox{65}{TextVQA} & \rotatebox{65}{VizWiz} & & \rotatebox{65}{\textbf{Avg}} & \rotatebox{65}{MMVP} & \rotatebox{65}{RWQA} & \rotatebox{65}{Blink} & \rotatebox{65}{V*} & \rotatebox{65}{CVBench} \\
\midrule
Qwen2.5-3B & & 73.5 & 63.8 & 87.7 & 1484.0 & 70.9 & 71.2 & & 46.7 & 75.4 & 40.9 & 7.7 & 62.9 & & 42.2 & 17.3 & 37.1 & 60.6 & 53.8 & & 50.5 & 46.0 & 55.2 & 46.8 & 44.5 & \textbf{60.2} \\
\rowcolor{OurColor}
\textbf{+ \ours (Ours)} & & \textbf{74.5} & \textbf{64.1} & \textbf{87.9} & \textbf{1525.5} & \textbf{72.0} & \textbf{72.0} & & \textbf{47.5} & \textbf{75.5} & \textbf{41.3} & \textbf{7.9} & \textbf{65.1} & & \textbf{43.8} & \textbf{20.2} & \textbf{39.1} & \textbf{61.7} & \textbf{54.3} & & \textbf{52.9} & \textbf{48.0} & \textbf{56.3} & \textbf{49.1} & \textbf{51.3} & 59.6 \\
\midrule
Qwen3-4B & & 75.6 & 64.6 & 87.3 & 1528.4 & 75.4 & 74.1 & & 49.6 & 77.2 & 44.6 & \textbf{8.4} & 68.4 & & 43.8 & 23.3 & \textbf{42.5} & \textbf{65.6} & 44.0 & & 56.3 & 48.0 & 59.9 & \textbf{55.1} & 50.3 & 68.3 \\
\rowcolor{OurColor}
\textbf{+ \ours (Ours)} & & \textbf{76.0} & \textbf{64.9} & \textbf{87.9} & \textbf{1544.6} & \textbf{75.6} & \textbf{74.3} & & \textbf{50.1} & \textbf{78.8} & \textbf{44.7} & 8.1 & \textbf{68.9} & & \textbf{44.5} & \textbf{24.8} & 40.5 & \textbf{65.6} & \textbf{47.2} & & \textbf{58.5} & \textbf{56.0} & \textbf{61.3} & 53.7 & \textbf{52.4} & \textbf{69.1} \\
\midrule
Vicuna-7B & & \textbf{72.2} & 64.9 & 87.5 & \textbf{1469.0} & \textbf{64.6} & 70.8 & & 44.3 & 70.3 & \textbf{36.3} & 11.1 & 59.4 & & \textbf{45.7} & 20.0 & \textbf{40.6} & 64.3 & \textbf{58.0} & & 49.7 & 38.7 & 56.5 & 46.8 & 44.5 & \textbf{62.1} \\
\rowcolor{OurColor}
\textbf{+ \ours (Ours)} & & 72.1 & \textbf{65.1} & \textbf{87.8} & 1453.4 & 64.0 & \textbf{70.9} & & \textbf{44.5} & \textbf{70.9} & 35.0 & \textbf{12.4} & \textbf{59.5} & & \textbf{45.7} & \textbf{20.6} & 40.3 & \textbf{64.4} & 57.6 & & \textbf{52.0} & \textbf{42.7} & \textbf{57.8} & \textbf{47.9} & \textbf{49.7} & 61.9 \\
\midrule
LLama3-8B & & 73.3 & 64.8 & 87.5 & \textbf{1506.0} & 67.4 & 71.5 & & 45.0 & 72.5 & 37.9 & 7.2 & 62.5 & & 43.0 & 18.8 & \textbf{40.1} & 63.8 & 49.5 & & 53.8 & 46.0 & 60.1 & 49.2 & 44.0 & \textbf{69.5} \\
\rowcolor{OurColor}
\textbf{+ \ours (Ours)} & & \textbf{74.6} & \textbf{65.6} & \textbf{87.6} & 1503.3 & \textbf{71.7} & \textbf{72.7} & & \textbf{46.3} & \textbf{75.9} & \textbf{38.0} & \textbf{8.4} & \textbf{63.1} & & \textbf{44.7} & \textbf{19.9} & 39.9 & \textbf{64.6} & \textbf{54.2} & & \textbf{55.1} & \textbf{46.7} & \textbf{60.4} & \textbf{50.2} & \textbf{51.8} & 66.4 \\
\midrule
Qwen2.5-7B & & 76.2 & 64.9 & 88.2 & \textbf{1582.2} & 75.3 & 73.7 & & 50.2 & \textbf{77.9} & 45.1 & \textbf{8.8} & 68.8 & & 47.9 & 24.2 & 39.9 & 64.9 & 62.5 & & 56.7 & 51.3 & 59.6 & \textbf{51.7} & \textbf{50.8} & 70.2 \\
\rowcolor{OurColor}
\textbf{+ \ours (Ours)} & & \textbf{76.5} & \textbf{65.3} & \textbf{88.3} & 1574.8 & \textbf{76.0} & \textbf{74.1} & & \textbf{50.5} & \textbf{77.9} & \textbf{46.9} & 7.6 & \textbf{69.6} & & \textbf{48.6} & \textbf{23.9} & \textbf{40.7} & \textbf{65.6} & \textbf{64.1} & & \textbf{57.4} & \textbf{54.0} & \textbf{61.3} & 50.2 & 50.3 & \textbf{71.3} \\
\midrule
Qwen3-8B & & 74.7 & 64.8 & 86.5 & 1472.8 & 75.4 & 73.1 & & 49.7 & 77.5 & \textbf{47.0} & 7.3 & 67.2 & & 43.5 & 20.2 & 37.4 & 64.3 & 51.9 & & 55.9 & 49.3 & 59.2 & 52.2 & \textbf{50.8} & 67.8 \\
\rowcolor{OurColor}
\textbf{+ \ours (Ours)} & & \textbf{76.9} & \textbf{66.1} & \textbf{87.6} & \textbf{1562.3} & \textbf{77.6} & \textbf{74.9} & & \textbf{51.1} & \textbf{78.8} & 46.2 & \textbf{8.6} & \textbf{70.8} & & \textbf{47.6} & \textbf{27.6} & \textbf{41.4} & \textbf{67.9} & \textbf{53.6} & & \textbf{59.5} & \textbf{58.0} & \textbf{60.1} & \textbf{55.5} & 50.3 & \textbf{73.5} \\
\midrule
Vicuna-13B & & 73.4 & 65.5 & 87.7 & 1491.9 & 67.1 & \textbf{72.2} & & 45.5 & 72.0 & \textbf{38.0} & 9.9 & \textbf{62.0} & & \textbf{47.7} & \textbf{22.2} & 41.6 & \textbf{67.2} & \textbf{59.5} & & 52.7 & \textbf{44.7} & \textbf{58.7} & 50.6 & 46.1 & 63.7 \\
\rowcolor{OurColor}
\textbf{+ \ours (Ours)} & & \textbf{73.6} & \textbf{65.8} & \textbf{88.0} & \textbf{1504.7} & \textbf{67.2} & 71.9 & & \textbf{45.7} & \textbf{72.2} & 36.6 & \textbf{13.2} & 60.8 & & 47.6 & 22.0 & \textbf{41.8} & 67.1 & 59.4 & & \textbf{53.9} & 44.0 & 57.3 & \textbf{52.3} & \textbf{49.2} & \textbf{66.5} \\
\midrule
Qwen2.5-14B & & 75.6 & 64.2 & \textbf{88.0} & 1548.8 & 75.5 & 72.7 & & 50.7 & 76.4 & 48.2 & 8.3 & 69.7 & & 44.8 & 19.6 & 33.9 & 63.7 & 61.9 & & 54.9 & 47.3 & \textbf{60.1} & 52.7 & 46.6 & 67.9 \\
\rowcolor{OurColor}
\textbf{+ \ours (Ours)} & & \textbf{77.4} & \textbf{66.1} & \textbf{88.0} & \textbf{1600.9} & \textbf{77.5} & \textbf{75.5} & & \textbf{52.8} & \textbf{78.6} & \textbf{49.1} & \textbf{9.6} & \textbf{74.1} & & \textbf{49.3} & \textbf{27.4} & \textbf{39.0} & \textbf{67.9} & \textbf{62.9} & & \textbf{58.3} & \textbf{52.0} & \textbf{60.1} & \textbf{55.2} & \textbf{52.4} & \textbf{71.6} \\
\midrule
Qwen3-14B & & 77.4 & 65.6 & 87.6 & 1609.5 & \textbf{78.5} & 74.6 & & \textbf{52.8} & \textbf{79.6} & \textbf{49.6} & \textbf{10.2} & 71.9 & & 46.1 & 26.8 & 41.1 & 69.7 & 47.0 & & 58.2 & 57.3 & 60.3 & \textbf{52.6} & 50.3 & \textbf{70.8} \\
\rowcolor{OurColor}
\textbf{+ \ours (Ours)} & & \textbf{77.7} & \textbf{66.3} & \textbf{88.0} & \textbf{1632.5} & 78.0 & \textbf{74.9} & & 52.6 & 79.4 & 49.1 & 9.2 & \textbf{72.6} & & \textbf{47.8} & \textbf{28.1} & \textbf{41.8} & \textbf{69.9} & \textbf{51.4} & & \textbf{58.9} & \textbf{58.0} & \textbf{62.5} & 52.0 & \textbf{51.8} & 70.2 \\
\bottomrule
\end{tabular}
}
\vspace{-0.2cm}
\end{table}

\begin{table}[t]
\caption{Detailed VQA results of different representation alignment strategies for MLLMs.}
\label{tab:comparison_full}
\vspace{-0.1cm}
\centering
\setlength{\tabcolsep}{0.25em}
\resizebox{\linewidth}{!}{
\begin{tabular}{lc cccccc c ccccc c ccccc c >{\columncolor{OurColor!30}}c>{\columncolor{OurColor!30}}c>{\columncolor{OurColor!30}}c>{\columncolor{OurColor!30}}c>{\columncolor{OurColor!30}}c>{\columncolor{OurColor!30}}c}
\toprule
& & \multicolumn{6}{c}{\textbf{General}}
& & \multicolumn{5}{c}{\textbf{Knowledge}}
& & \multicolumn{5}{c}{\textbf{OCR}}
& & \multicolumn{6}{c}{\cellcolor{OurColor!30}\textbf{Vision}} \\
\cmidrule{3-8}
\cmidrule{10-14}
\cmidrule{16-20}
\cmidrule{22-27}
\textbf{Alignment} & & \rotatebox{65}{\textbf{Avg}} & \rotatebox{65}{GQA} & \rotatebox{65}{POPE} & \rotatebox{65}{MME} & \rotatebox{65}{MMB} & \rotatebox{65}{SEED} & & \rotatebox{65}{\textbf{Avg}} & \rotatebox{65}{SQA} & \rotatebox{65}{MMMU} & \rotatebox{65}{MathV} & \rotatebox{65}{AI2D} & & \rotatebox{65}{\textbf{Avg}} & \rotatebox{65}{ChartQA} & \rotatebox{65}{OCRB} & \rotatebox{65}{TextVQA} & \rotatebox{65}{VizWiz} & & \rotatebox{65}{\textbf{Avg}} & \rotatebox{65}{RWQA} & \rotatebox{65}{MMVP} & \rotatebox{65}{Blink} & \rotatebox{65}{V*} & \rotatebox{65}{CVBench} \\
\midrule
- & & 74.7 & 64.8 & 86.5 & 1472.7 & 75.4 & 73.1 & & 49.7 & 77.5 & \textbf{47.0} & 7.3 & 67.2 & & 43.5 & 20.2 & 37.4 & 64.3 & 51.9 & & 55.9 & 59.2 & 49.3 & 52.2 & 50.8 & 67.8 \\
ROSS~\cite{wangreconstructive} & & 74.6 & 64.5 & 87.2 & 1472.5 & 74.9 & 72.7 & & 49.4 & 76.4 & 46.3 & 7.4 & 67.4 & & 44.0 & 17.8 & 36.8 & 64.8 & \textbf{56.8} & & 56.2 & 59.8 & 50.3 & 52.2 & 49.2 & 69.7 \\
VIRAL~\cite{yoon2025visual} & & 73.8 & 64.2 & 87.3 & 1396.6 & 74.9 & 72.6 & & 49.3 & 76.4 & 45.7 & 8.0 & 67.1 & & 43.6 & 19.3 & 36.7 & 64.2 & 54.2 & & 54.2 & 57.6 & 46.7 & 51.3 & 46.1 & 69.3 \\
JARVIS~\cite{caffagni2025seeing} & & 76.8 & 64.6 & \textbf{88.2} & \textbf{1605.1} & 76.7 & 74.2 & & 49.9 & 77.0 & 45.4 & 8.4 & 68.6 & & 46.2 & 27.1 & 40.5 & 66.0 & 51.2 & & 58.7 & 59.2 & 54.7 & 54.2 & \textbf{55.5} & 69.9 \\
CMAR~\cite{gan2025cross} & & 76.4 & 65.6 & 87.4 & 1556.6 & 76.5 & 74.7 & & 51.0 & 78.3 & 45.9 & \textbf{8.6} & \textbf{71.0} & & 46.0 & 23.3 & \textbf{41.6} & 67.6 & 51.6 & & 56.9 & 58.8 & 50.0 & 52.1 & 52.9 & 70.9 \\
\rowcolor{OurColor}
\textbf{\ours (Ours)} & & \textbf{76.9} & \textbf{66.1} & 87.6 & 1562.3 & \textbf{77.6} & \textbf{74.9} & & \textbf{51.1} & \textbf{78.8} & 46.2 & \textbf{8.6} & 70.8 & & \textbf{47.6} & \textbf{27.6} & 41.4 & \textbf{67.9} & 53.6 & & \textbf{59.5} & \textbf{60.1} & \textbf{58.0} & \textbf{55.5} & 50.3 & \textbf{73.5} \\
\bottomrule
\end{tabular}
}\vspace{-0.25cm}
\end{table}

\tit{Extended Results on All Benchmarks} As we aim to improve MLLMs, we focus on strengthening their visual perception, which is particularly stressed on vision-centric benchmarks. Consequently, in the main paper, we reported detailed scores on specific vision-centric VQA datasets, such as RealWorldQA, MMVP, Blink, V*, and CVBench, leaving the average score on the General, Knowledge, and OCR categories. Here, we report the full results over the 18 VQA benchmarks considered in our study. Specifically, we refer to Table~\ref{tab:llms_full} for the results of different LLM families, and to Table~\ref{tab:comparison_full} for a detailed comparison between representation alignment methods on the Qwen3-8B LLM.

\begin{figure}
    \centering
    \includegraphics[width=\linewidth]{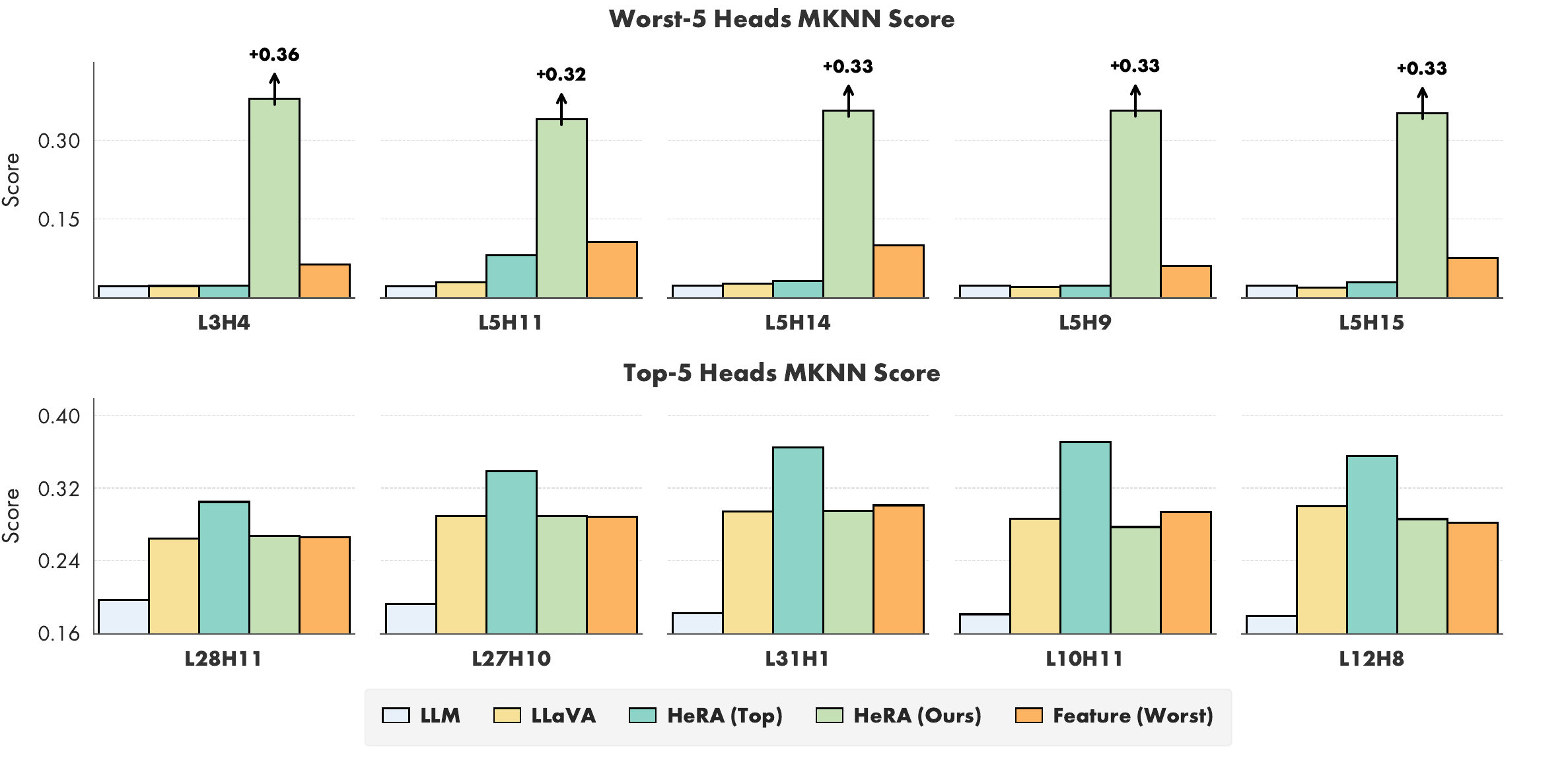}
    \vspace{-0.75cm}
    \caption{Effect of the multimodal training of LLaVA and representation alignment methods on the Worst-5 and Top-5 heads. Worst-5 and Top-5 heads are selected by the lowest and highest MKNN alignment score with DINOv2-L, computed on the Qwen2.5-3B LLM before multimodal training.}
    \label{fig:mknn_top_supp}
    \vspace{-0.3cm}
\end{figure}

\subsection{Additional Analyses}

\tit{Additional MKNN Head-Wise Analysis} In Fig.~\ref{fig:mknn_top_supp}, we provide a detailed comparison of the MKNN alignment scores for the Worst-5 (\textit{upper} half) and Top-5 (\textit{bottom} half) attention heads across different training strategies. These specific heads are identified by computing their MKNN alignment scores on the base Qwen2.5-3B LLM prior to any multimodal training. Interestingly, we observe that the relative alignment of these heads is largely preserved after standard multimodal training (\ie, LLaVA): heads that are naturally highly aligned in the base LLM remain highly aligned, whereas poorly aligned heads stay poorly aligned.

When we apply \ours to the Top-5 heads, their alignment scores further increase, but this intervention has absolutely no impact on the Worst-5 heads. As shown in Tab.~\ref{tab:ablation} (\textit{seventh} row), this translates into suboptimal downstream performance. Conversely, our proposed strategy, that is \ours applied to the Worst-5 heads, greatly increases the alignment of the targeted components. Crucially, this massive boost does not sacrifice the integrity of the Top-5 heads, which record MKNN alignment scores remarkably similar to the LLaVA baseline.

Finally, we observe that enforcing representation alignment at the feature level, specifically, via cosine similarity maximization between the MLLM visual features and the teacher vision encoder, is ineffective at modifying the local topological structure (nor at improving performance, see Table~\ref{tab:ablation}, \textit{third} row). As depicted in the plot, this approach has a very small effect on the MKNN alignment scores of the targeted Worst-5 heads, further highlighting the unique contribution of our topology-aware contrastive objective.

\begin{figure}[t]
    \centering
    \includegraphics[width=\linewidth]{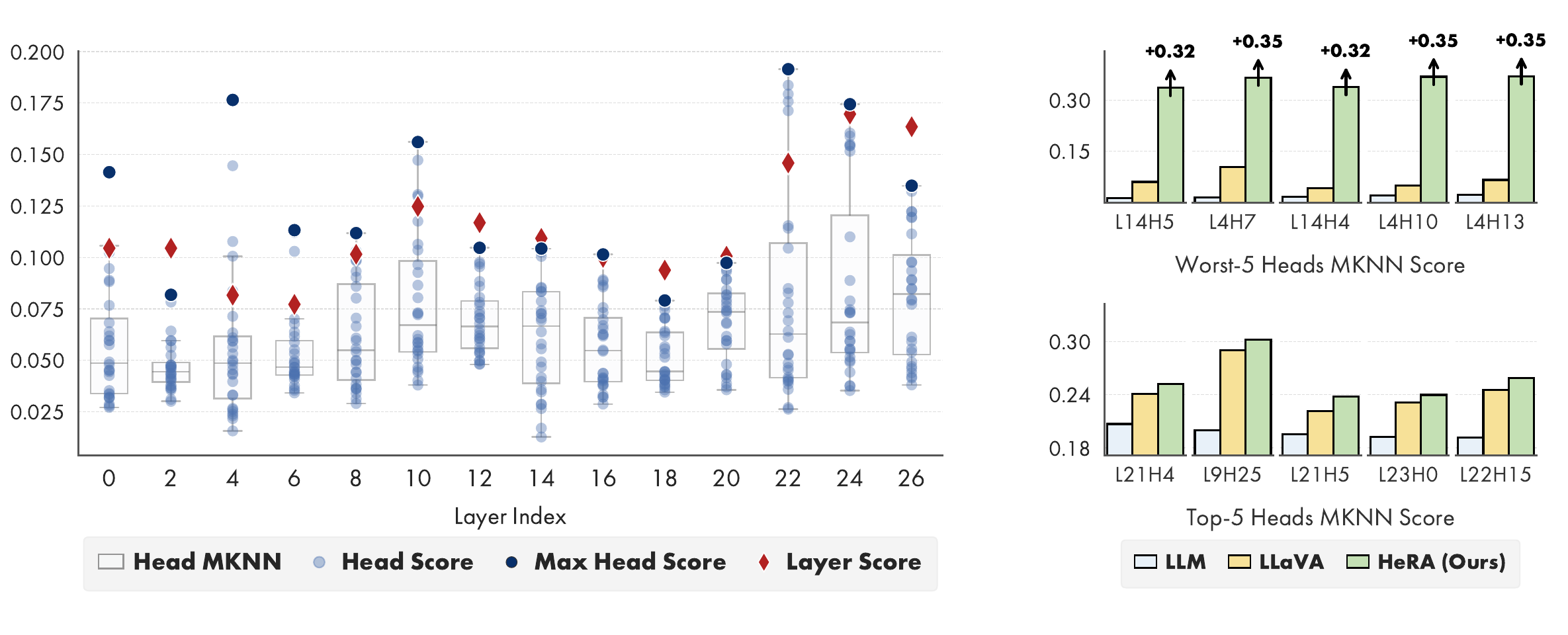}
    \vspace{-0.75cm}
    \caption{\textbf{Left:} Alignment with DINOv2-L, measured with the MKNN metric on each layer and attention head of Qwen2.5-7B. \textbf{Right:} MKNN scores of the Worst-5 and Top-5 heads, computed on \textit{(i)} the base LLM; \textit{(ii)} after the LLaVA multimodal training; and \textit{(iii)} after the addition of \ours.}
    \label{fig:mknn_qwen2_5_7B}
    \vspace{-0.3cm}
\end{figure}

\begin{figure}[t]
    \centering
    \includegraphics[width=\linewidth]{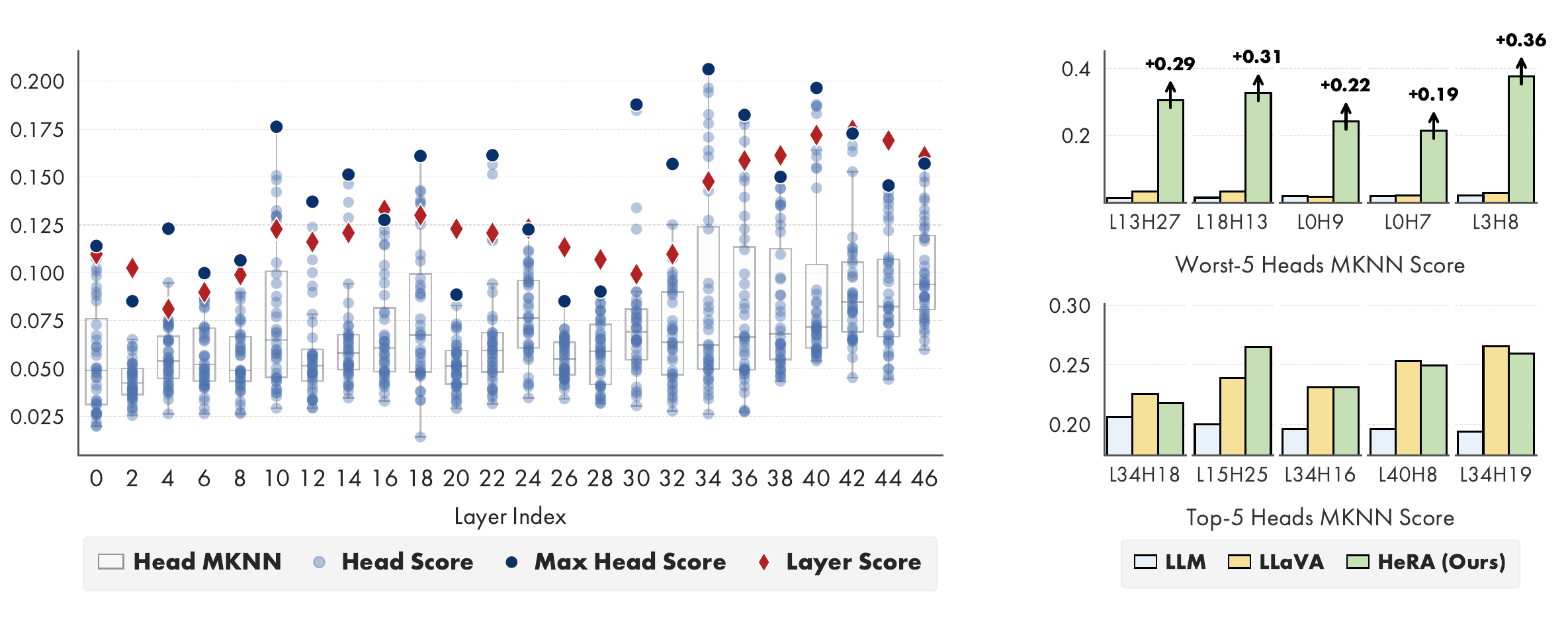}
    \vspace{-0.75cm}
    \caption{\textbf{Left:} Alignment with DINOv2-L, measured with the MKNN metric on each layer and attention head of Qwen2.5-14B. \textbf{Right:} MKNN scores of the Worst-5 and Top-5 heads, computed on \textit{(i)} the base LLM; \textit{(ii)} after the LLaVA multimodal training; and \textit{(iii)} after the addition of \ours.}
    \label{fig:mknn_qwen2_5_14B}
    \vspace{-0.3cm}
\end{figure}

\tit{MKNN Alignment With Larger Qwen2.5 Models} In Fig.~\ref{fig:mknn_qwen2_5_7B} and Fig.~\ref{fig:mknn_qwen2_5_14B}, we extend the MKNN alignment analysis of Fig.~\ref{fig:mknn_plot} to larger LLMs within the same family, specifically evaluating Qwen2.5-7B and Qwen2.5-14B against the DINOv2-L teacher. The left parts confirm that our previous observations persist at larger scales: representations of specific individual attention heads consistently exhibit a much higher natural alignment with the visual domain than those obtained out of any layer in the same model. Furthermore, the right parts highlight the atomic nature of our intervention. Applying \ours to the Worst-5 heads successfully drives a massive boost in their cross-modal alignment, without disrupting the structural alignment of the Top-5 heads that are already naturally aligned in the base LLM.

\begin{figure}[t]
    \centering
    \includegraphics[width=\linewidth]{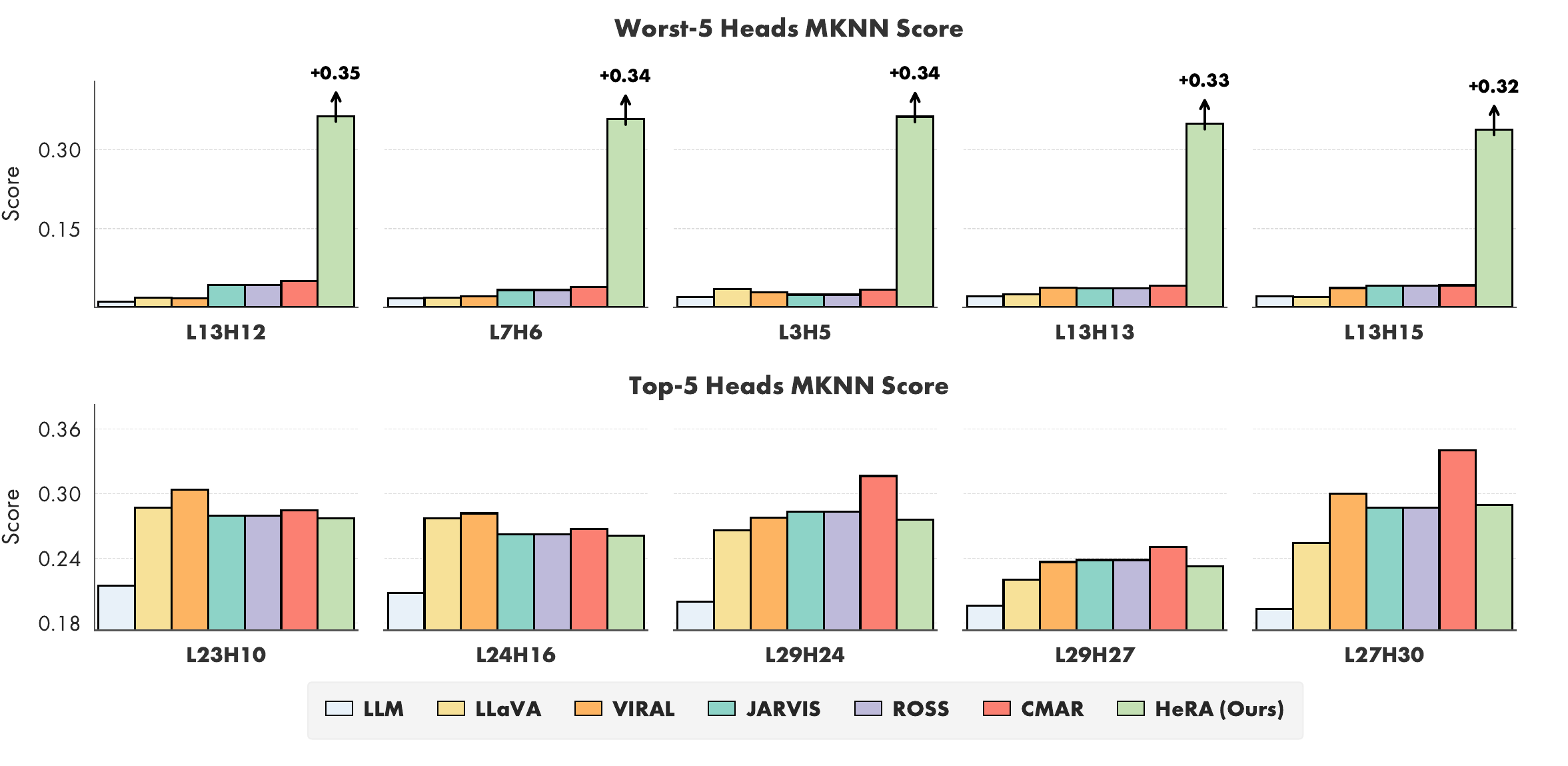}
    \vspace{-0.6cm}
    \caption{Comparison of MKNN scores of the Worst-5 and Top-5 heads after the second training stage performed with \ours and our competitors (\ie \space LLaVA, VIRAL, JARVIS, ROSS, CMAR).}
    \label{fig:mknn_competitor_supp}
    \vspace{-0.2cm}
\end{figure}

\tit{MKNN Analysis with Different Representation Alignment Methods} In Fig.~\ref{fig:mknn_competitor_supp}, we compare the MKNN alignment scores of \ours against existing representation alignment strategies. For this analysis, all methods are trained using Qwen3-8B as the LLM and SigLIP2 as the vision encoder, with the MKNN alignment metric computed with respect to the DINOv2-L teacher. We remind to Table~\ref{tab:comparison} for a quantitative comparison on VQA benchmarks.

Consistent with our previous findings, all methods increase the alignment of the Top-5 heads. However, this is largely a natural consequence of the multimodal training process itself, as most methods achieve scores that closely mirror the unregularized LLaVA baseline. The only method that registers a more significant, distinct impact on the Top-5 heads is CMAR. CMAR shares a conceptual similarity with \ours in that it enforces cross-modal topological alignment rather than strict feature-level visual matching. However, a key difference lies in their scope: CMAR relies on the CKA metric to match \textit{global} pairwise relationships across all samples within a training batch, whereas \ours strictly targets the consistency of \textit{local} neighborhoods. 

Regarding the analysis on the Worst-5 heads, both CMAR and feature-matching methods fail to induce any meaningful structural changes. Conversely, \ours is the \textit{only} method capable of significantly increasing the cross-modal alignment of these initially poorly aligned heads.

\begin{figure}[t]
    \centering
    \includegraphics[width=0.8\linewidth]{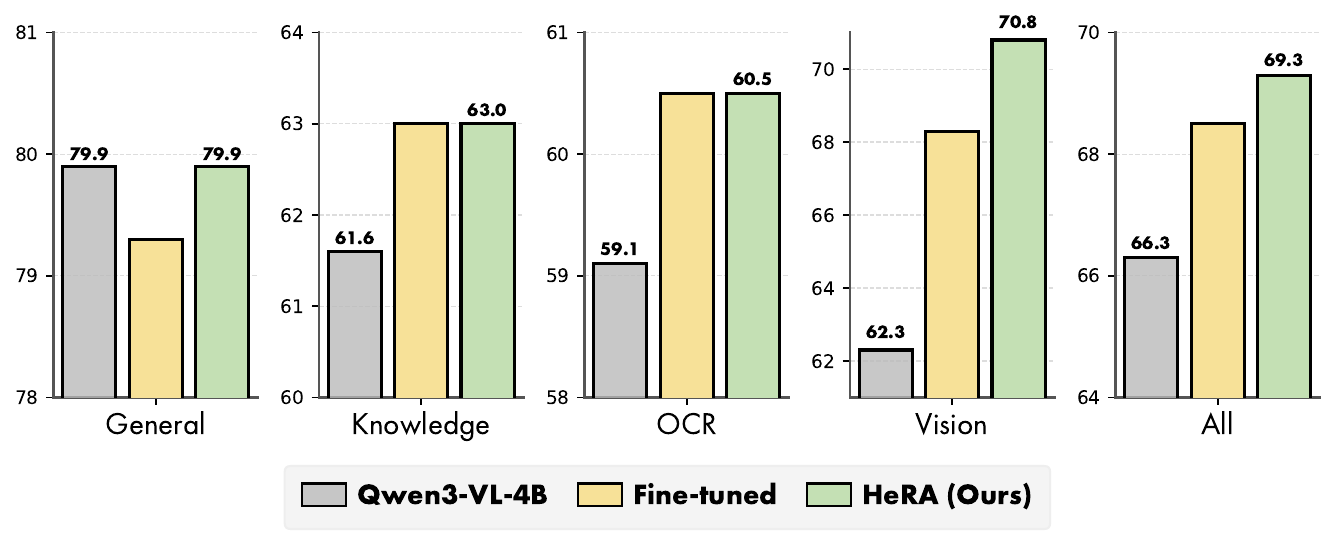}
    \vspace{-0.3cm}
    \caption{VQA results on Qwen3-VL-4B after fine-tuning, with and without the \ours objective.}
    \label{fig:vlm_supp}
    \vspace{-0.3cm}
\end{figure}

\subsection{Qualitative Results}
\label{sec:appendix_qualitatives}
\begin{figure}[t]
    \centering
    \includegraphics[width=0.98\linewidth]{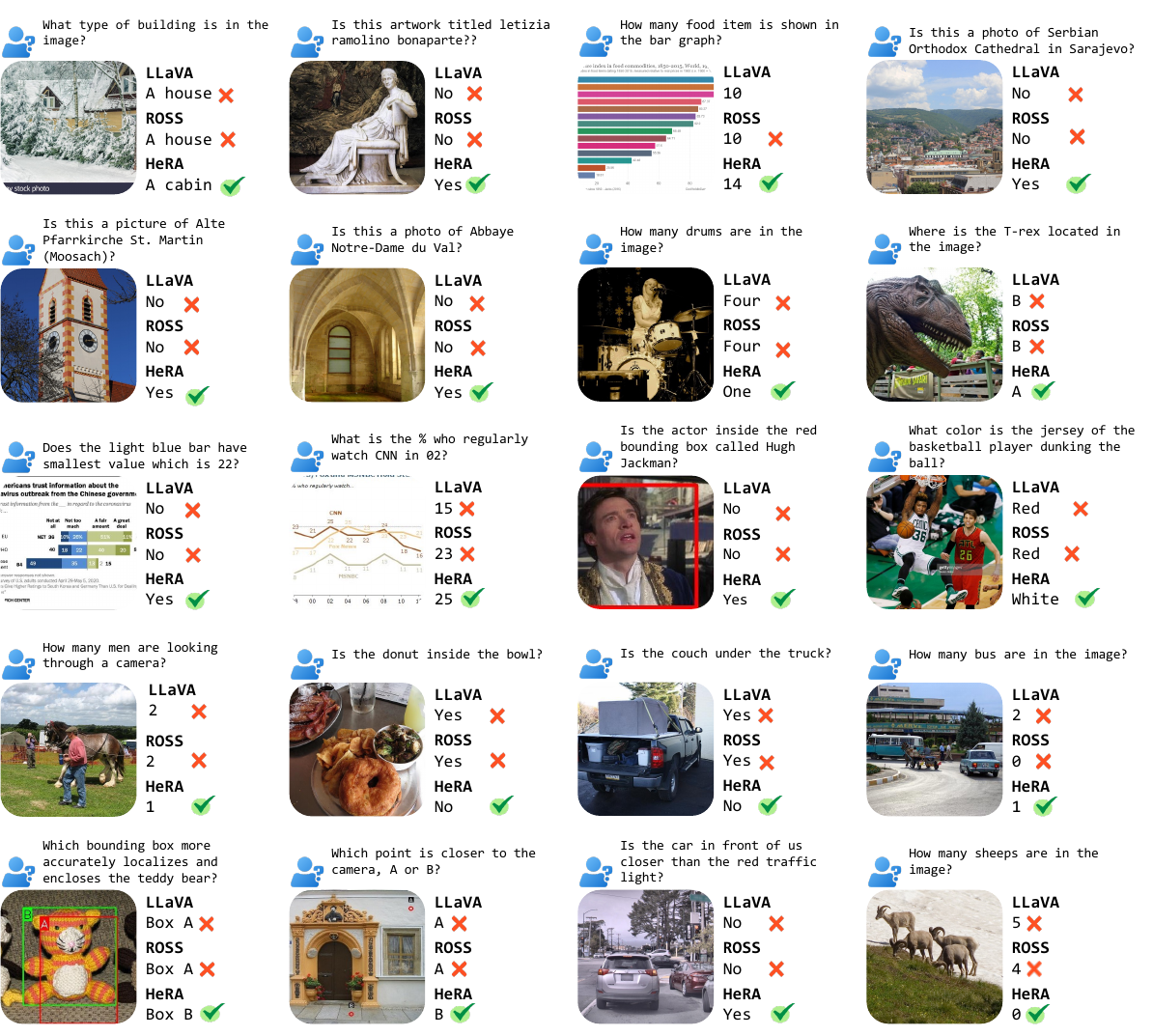}
    \vspace{-0.2cm}
    \caption{Qualitative comparison of LLaVA~\cite{liu2024improved}, ROSS~\cite{wangreconstructive}, and \ours using Qwen3-8B and SigLIP2. We present representative samples across all Cambrian categories: General, Knowledge, OCR, and Vision-Centric.}    
    \label{fig:qual_hera}
    \vspace{-0.35cm}
\end{figure}

In Fig.~\ref{fig:qual_hera}, we provide a qualitative comparison between the LLaVA~\cite{liu2024improved} baseline, ROSS~\cite{wangreconstructive}, and \ours using Qwen3-8B and SigLIP2. The representative samples demonstrate that \ours consistently delivers more accurate and better-grounded answers across all evaluated categories (General, Knowledge, OCR, and Vision-Centric), effectively correcting various perceptual errors made by the baselines. 

\begin{figure}[t]
    \centering
    \includegraphics[width=0.98\linewidth]{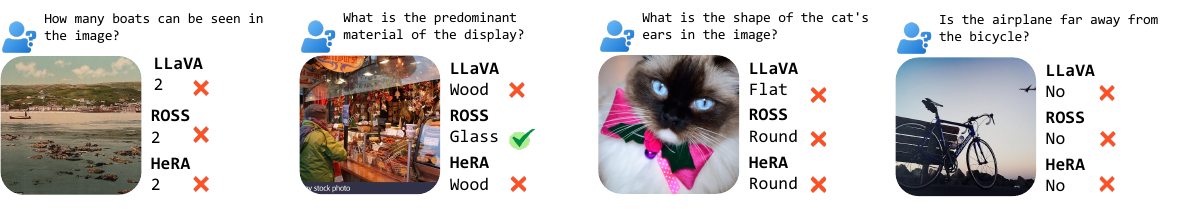}
    \vspace{-0.2cm}
    \caption{Failure cases of \ours on VQA tasks.}    
    \label{fig:qual_hera_2}
    \vspace{-0.35cm}
\end{figure}

Despite these clear improvements, in Fig.~\ref{fig:qual_hera_2}, we report a few failure cases where our model still struggles. Specifically, \ours can occasionally misinterpret fine-grained visual details, such as accurately counting multiple small instances, identifying ambiguous materials and shapes, or inferring precise spatial relationships in cluttered scenes.

\section{Limitations and Societal Impacts}
\label{sec:appendix_limitations}
We are aware that the landscape of MLLMs has rapidly evolved beyond LLaVA with the introduction of frontier proprietary models (as acknowledged at the end of Sec.~\ref{sec:related}). However, we adopted the LLaVA pipeline because it remains computationally tractable, allowing for the extensive ablation studies and rigorous evaluations presented in this work.

To bridge this gap and explore the potential of our method on modern architectures, we conducted a pioneering study applying \ours directly to Qwen3-VL-4B~\cite{bai2025qwen3}, a state-of-the-art multimodal LLM. We fine-tuned the model using an 83k-sample Cambrian split derived from the FineVision~\cite{wiedmann2025finevision} dataset. As shown in Fig.~\ref{fig:vlm_supp}, \ours records promising results, particularly on demanding vision-centric tasks. Crucially, the improvements on visual benchmarks are substantially higher with \ours than with standard fine-tuning alone; moreover, simple fine-tuning actually registers a performance regression in the General category, which does not manifest in \ours.

Beyond this architectural constraint, we do not foresee direct negative societal impacts arising specifically from our representation alignment technique. Rather, by improving visual grounding and mitigating object hallucinations, \ours contributes to the development of more reliable and factual vision-language systems.

\end{document}